\pdfoutput=1

\documentclass[11pt]{article}

\usepackage[preprint]{acl}

\usepackage{times}
\usepackage{latexsym}

\usepackage[T1]{fontenc}

\usepackage[utf8]{inputenc}

\usepackage{microtype}

\usepackage{inconsolata}

\usepackage{graphicx}
\usepackage{amsmath}
\usepackage[linesnumbered,ruled,vlined]{algorithm2e}
\usepackage{amssymb}
\usepackage{mathtools}
\usepackage{amsthm}
\usepackage{hyperref}
\usepackage{amssymb}
\usepackage{multirow}
\usepackage{xcolor}
\usepackage{colortbl}
\usepackage{longtable}  %
\usepackage{booktabs}

\usepackage{algorithmicx}
\usepackage{algpseudocode}
\usepackage{verbatim}
\usepackage{caption}
\usepackage{subcaption}
\usepackage{fancyvrb}
\usepackage{fvextra}

\definecolor{GREEN}{HTML}{62b197}
\definecolor{RED}{HTML}{e18e6d}

\DeclareUnicodeCharacter{FF5C}{\textbar}

\definecolor{graybg}{gray}{0.95}

\newcommand{\best}[1]{\textbf{#1}}
\newcommand{\secbest}[1]{\underline{#1}}

\usepackage{tcolorbox}
\tcbuselibrary{breakable}
\tcbuselibrary{breakable, skins, listings} 
\usepackage{cuted} 
\usepackage{float}
\usepackage{multicol}
\usepackage{enumitem}
\usepackage{pifont}
\usepackage{array}
\usepackage{listings}

\usepackage{geometry}
\definecolor{cgreen}{RGB}{0, 150, 0}
\definecolor{cred}{RGB}{200, 0, 0}
\definecolor{lightgray}{RGB}{245, 245, 245}

\usepackage{fontawesome5}

\definecolor{userframe}{RGB}{44, 62, 80}
\definecolor{usercolor}{RGB}{248, 250, 252}
\definecolor{modelframe}{RGB}{0, 128, 128}
\definecolor{modelcolor}{RGB}{240, 255, 255}
\definecolor{toolframe}{RGB}{211, 84, 0}
\definecolor{toolcolor}{RGB}{253, 242, 233}
\definecolor{agentframe}{RGB}{113, 54, 138} 
\definecolor{agentcolor}{RGB}{250, 245, 255}

\lstdefinestyle{jsonstyle}{
    basicstyle=\ttfamily\small,
    breaklines=true,         %
    breakatwhitespace=false, %
    postbreak=\mbox{\textcolor{red}{$\hookrightarrow$}\space}, %
    frame=none,
    backgroundcolor=\color{toolcolor}, %
    columns=fullflexible,    %
    keepspaces=true,
    showstringspaces=false,
    aboveskip=0pt,           %
    belowskip=0pt            %
}

\newtcolorbox{userbox}[1][]{
    enhanced,                %
    breakable,               %
    colback=usercolor,
    colframe=userframe,
    title={\ifdefined\faUser\faUser\ \fi\textbf{User}}, %
    fonttitle=\bfseries,
    arc=3mm,
    boxrule=0.8pt,
    left=2mm, right=2mm, top=2mm, bottom=2mm, %
    before skip=10pt,        %
    after skip=10pt,         %
    #1                       %
}

\newcommand{\profileitem}[3]{%
    \makebox[2.0em][c]{\textcolor{userframe}{#1}}%
    \textbf{#2:} \textcolor{darkgray}{#3} \\[0.4em]%
}

\newtcolorbox{modelbox}[1][]{
    enhanced,
    breakable,
    colback=modelcolor,
    colframe=modelframe,
    title={\ifdefined\faRobot\faRobot\ \fi\textbf{Model}},
    fonttitle=\bfseries,
    arc=3mm,
    boxrule=0.8pt,
    left=2mm, right=2mm, top=2mm, bottom=2mm,
    before skip=10pt,
    after skip=10pt,
    #1
}

\newtcolorbox{toolbox}[1][]{
    enhanced,
    breakable,
    colback=toolcolor,
    colframe=toolframe,
    title={\ifdefined\faTools\faTools\ \fi\textbf{Tool Call / Output}},
    fonttitle=\bfseries\small,
    fontupper=\small\ttfamily, %
    arc=1mm,
    boxrule=0.8pt,
    left=2mm, right=2mm, top=1mm, bottom=1mm,
    before skip=10pt,
    after skip=10pt,
    #1
}

\newtcolorbox{agentbox}[1][]{
    enhanced,
    breakable,
    colback=agentcolor,
    colframe=agentframe,
    title={\ifdefined\faUserSecret\faUserSecret\ \fi\textbf{Global Agent}}, %
    fonttitle=\bfseries,
    arc=3mm,
    boxrule=0.8pt,
    left=2mm, right=2mm, top=2mm, bottom=2mm,
    before skip=10pt,
    after skip=10pt,
    fontupper=\small, %
    #1
}

\newtcolorbox{promptbox}[1][]{
    enhanced,
    colback=gray!10,      
    colframe=gray!60,    
    coltitle=black,      
    fonttitle=\bfseries\large,
    title={#1},          
    sharp corners,        
    boxrule=1pt,          
    left=10pt, right=10pt, top=10pt, bottom=10pt, 
    fontupper=\small\ttfamily, 
}

\title{\centering FinToolSyn: A forward synthesis Framework for Financial Tool-Use Dialogue Data with Dynamic Tool Retrieval}

\author{
    \bf{\normalsize
    Caishuang Huang$^{1,2}$\thanks{Equal contributions.}, \ \ 
    Yang Qiao$^{2*}$, \ \ 
    Rongyu Zhang$^{3*}$, \ \ 
    Junjie Ye$^{1}$,} \ \ 
    Pu Lu$^{2}$, \\
    \bf{\normalsize 
    Wenxi Wu$^{2}$, \ \ 
    Meng Zhou$^{2}$,
    Xiku Du$^{2}$, \ \ 
    Tao Gui$^{1}$\thanks{Corresponding authors.}, \ \ 
    Qi Zhang$^{1}$\footnotemark[2], \ \ 
    Xuanjing Huang$^{1}$\footnotemark[2]} \\ 
  {$^1$ \normalsize College of Computer Science and Artificial Intelligence, Fudan University} \\
  {$^2$ \normalsize FiT, Tencent} \\
  {$^3$ \normalsize Zhejiang University}\\
  \texttt{\normalsize cshuang23@m.fudan.edu.cn, \{qz, tgui, xjhuang\}@fudan.edu.cn} \\
}

\begin{document}
\maketitle
\begin{abstract}
Tool-use capabilities are vital for Large Language Models (LLMs) in finance, a domain characterized by massive investment targets and data-intensive inquiries. However, existing data synthesis methods typically rely on a reverse synthesis paradigm, generating user queries from pre-sampled tools. This approach inevitably introduces artificial explicitness, yielding queries that fail to capture the implicit, event-driven nature of real-world needs. Moreover, its reliance on static tool sets overlooks the dynamic retrieval process required to navigate massive tool spaces. To address these challenges, we introduce \textit{FinToolSyn}, a forward synthesis framework designed to generate high-quality financial dialogues. Progressing from persona instruction and atomic tool synthesis to dynamic retrieval dialogue generation, our pipeline constructs a repository of 43,066 tools and synthesizes over 148k dialogue instances, incorporating dynamic retrieval to emulate the noisy candidate sets typical of massive tool spaces. We also establish a dedicated benchmark to evaluate tool-calling capabilities in realistic financial scenarios. Extensive experiments demonstrate that models trained on FinToolSyn achieve a 21.06\% improvement, providing a robust foundation for tool learning in financial scenarios.

\end{abstract}

\section{Introduction}

\begin{figure}[!t]
    \centering
    \includegraphics[width=\linewidth]{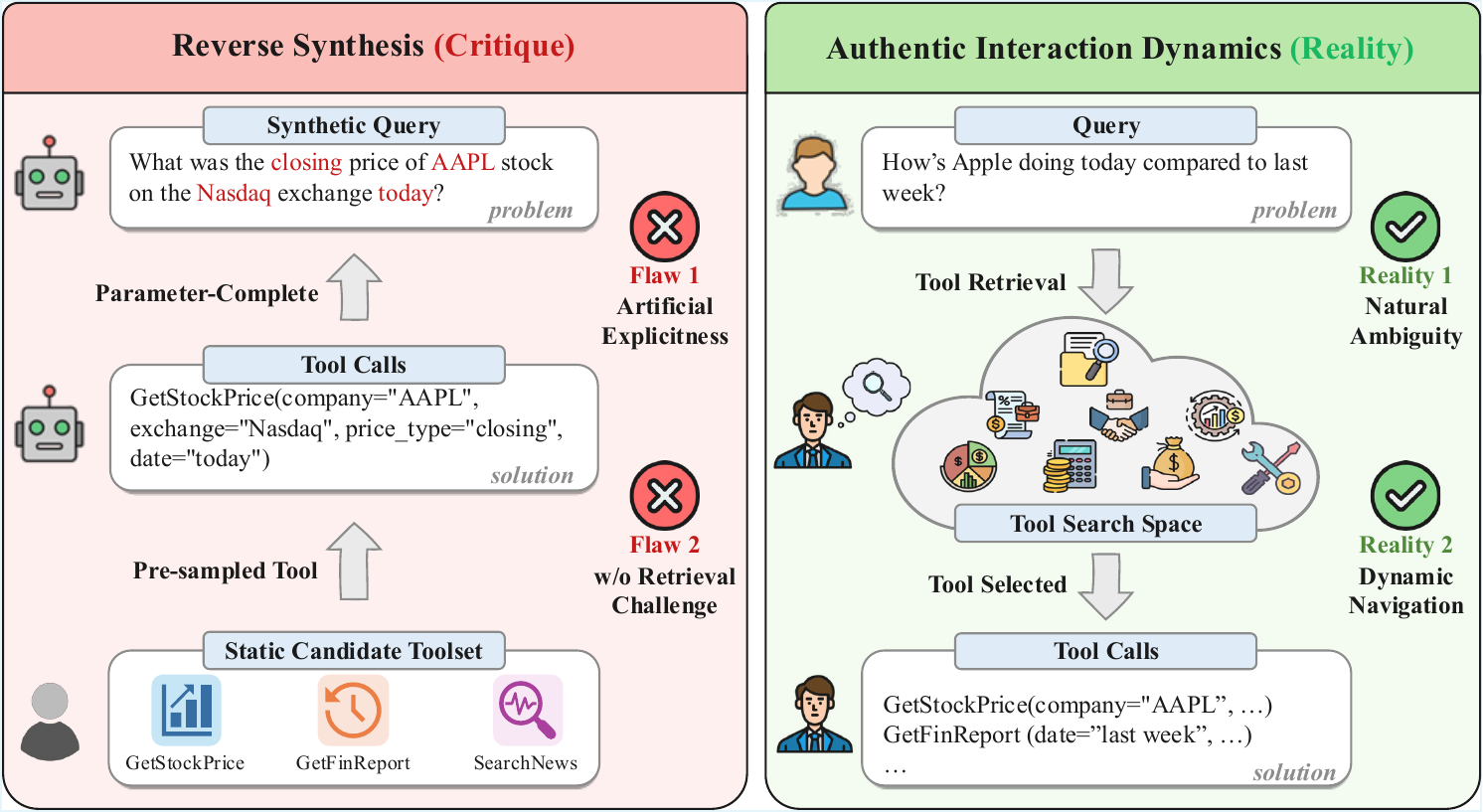} %
    \caption{Comparison between reverse synthesis and authentic interaction dynamics.}
    \vspace{-4mm} %
    \label{fig:head}
\end{figure}

Equipping Large Language Models (LLMs)~\cite{openai2024gpt4technicalreport,yang2025qwen3technicalreport,deepseekai2025deepseekv3technicalreport} with tool-use capabilities marks a critical evolution in the financial domain~\cite{Theuma2024EquippingLM,hu2025finsearchcomprealisticexpertlevelevaluation}. 
Given the sector's stringent demands for high timeliness and professional rigor, the static knowledge of pre-trained LLMs often proves insufficient~\cite{shah2025reportedcutofflargelanguage}. 
Consequently, invoking external tools is indispensable for executing complex, specialized tasks in this landscape.
However, the prevailing ``reverse synthesis'' paradigm, in which queries are generated conditional on pre-sampled tools, is fundamentally misaligned with authentic interaction dynamics.
As illustrated in Figure \ref{fig:head}, this approach introduces a logical shortcut where the ``solution'' artificially precedes the ``problem''~\cite{yang2025toolmindtechnicalreportlargescale,liu2024toolace}.
This results in data suffering from ``artificial explicitness'', where synthetic queries typically contain overly complete parameter information, thereby eliminating the natural ambiguity inherent in human communication.
Furthermore, reliance on static tool sets removes the challenge of retrieval; it fails to simulate the cognitive burden of dynamic navigation, where an agent must identify the correct utility within a massive, noisy search space rather than selecting from a limited candidate set~\cite{ye2024tooleyesfinegrainedevaluationtool,BFCL}.

To address these challenges, we propose \textit{FinToolSyn}, a ``forward synthesis'' framework designed to reconstruct high-quality, event-driven financial dialogues from the ground up. This pipeline follows an authentic demand trajectory, progressing from persona-driven instruction anchoring and demand-oriented atomic tool synthesis to multi-turn dialogue generation with dynamic retrieval.
To illustrate the difference, consider a traditional reverse-synthesized query derived from \texttt{get\_price(ticker=``AAPL'')}: \textit{``Tell me the price of Apple (AAPL).''} Here, the ticker is explicitly provided, reducing the task to trivial slot-filling. In contrast, our forward synthesis generates realistic queries such as: \textit{``Is my portfolio safe after the recent news about Buffett trimming his stake?''} — which is implicit, event-driven, and requires multi-step reasoning.
By integrating this demand-oriented synthesis with dynamic candidate tool retrieval, the framework naturally encapsulates real-world interaction dynamics and emulates the cognitive complexity of navigating massive, noisy tool spaces. This process ultimately yields a comprehensive repository of 43,066 customized atomic tools and over 148,000 dialogue instances, capturing the implicit, cognitive patterns of real-world financial inquiries.

Constructed upon this resource, we establish \textit{FinToolBench}, a dedicated benchmark comprising 843 human-verified gold-standard samples to rigorously evaluate tool-calling capabilities in realistic financial scenarios. To enforce financial safety in high-stakes contexts, we propose the Circuit-Breaker Hierarchical Weighted Scoring (CB-HWS) mechanism, which integrates ``fail-fast'' checks with hierarchical error penalization. Furthermore, we introduce two domain-specific metrics, Key Digit Accuracy (KDA) and Invocation Timing Accuracy (ITA), designed to enforce zero-tolerance precision on critical financial entities and validate strict pre-execution compliance.

Experiments on LLMs of varying scales demonstrate that fine-tuning on FinToolSyn yields substantial performance gains in specialized financial tool-use. Notably, domain-tuned models exhibit significant advantages in handling clarification inquiries, achieving superior capability in eliciting missing parameters critical for tool execution and adhering to logical business workflows. These results confirm that our framework effectively enhances contextual reasoning and decision-making within complex financial environments.

In summary, our main contributions are:
\begin{itemize} 
    \item We propose FinToolSyn, a comprehensive dialogue synthesis pipeline for financial tool learning, and establish FinToolBench, a specialized benchmark to rigorously assess model performance.
    \item We introduce a novel ``forward synthesis'' paradigm featuring persona-grounded instruction generation and task-decomposed tool construction, effectively overcoming the ``artificial explicitness'' bias in traditional ``reverse synthesis''.
    \item We incorporate globally-controlled dynamic retrieval to simulate navigation within massive tool spaces. Experiments validate that our method significantly enhances financial tool-use capabilities while maintaining strong generalization across general domains.
\end{itemize}

\section{Related Works}

\paragraph{Tool Calling for LLMs}
External tools enable LLMs to surpass static knowledge limits, excelling in reasoning~\cite{he2023solving,gou2023tora} and code execution~\cite{gao2023pal,zhang2024codeagent}. Beyond tuning-free ReAct~\cite{yao2022react}, tuning-based methods like ToolLLaMA~\cite{qin2023toolllm} refine tool-use via specialized datasets. However, their efficacy remains heavily contingent upon the quality and diversity of training corpora.

\paragraph{Synthetic Data for Tool Use}
To address data scarcity, recent work has turned to large-scale synthetic generation for tool-use training~\cite{xu2023tool,wang2025toolflow}. Most existing approaches adopt a reverse synthesis paradigm, generating queries conditional on pre-sampled tools, which introduces ``artificial explicitness'' bias and produces overly explicit requests that deviate from real-world information asymmetry~\cite{liu2024toolace,yang2025toolmindtechnicalreportlargescale}. In contrast, our forward synthesis framework grounds generation in latent user needs prior to tool exposure, better preserving the implicit nature of authentic interactions.

\paragraph{Candidate Tool Retrieval}
Efficient tool retrieval is critical under limited context budgets~\cite{shi2025retrieval}, yet existing methods struggle with complex, multi-step instructions in dynamic environments~\cite{qu2024towards,zheng2024toolrerank}. Moreover, benchmarks typically assume clean candidate sets, underestimating the challenges of realistic tool discovery and generalization~\cite{ye2024rotbenchmultilevelbenchmarkevaluating}. FinToolSyn introduces a globally controlled dynamic retrieval process to model noisy, exploratory tool selection over large, heterogeneous repositories.

\paragraph{LLMs in Finance}
While domain-specific models like BloombergGPT~\cite{wu2023bloomberggpt} and FinGPT~\cite{wang2023fingpt} have advanced financial NLP, high-quality tool-calling data remains scarce due to privacy barriers. Even recent efforts like FinMCP-Bench~\cite{finmcpbench2025}, which introduced multi-turn scenarios, are limited in scale. Consequently, existing models frequently suffer from reasoning failures in quantitative analysis due to a lack of logically coherent training corpora. We bridge this gap through demand-driven atomic tool synthesis, constructing a modular ecosystem that supports robust tool learning in vertical domains.

\begin{figure*}[!t]
    \centering
    \includegraphics[width=\linewidth]{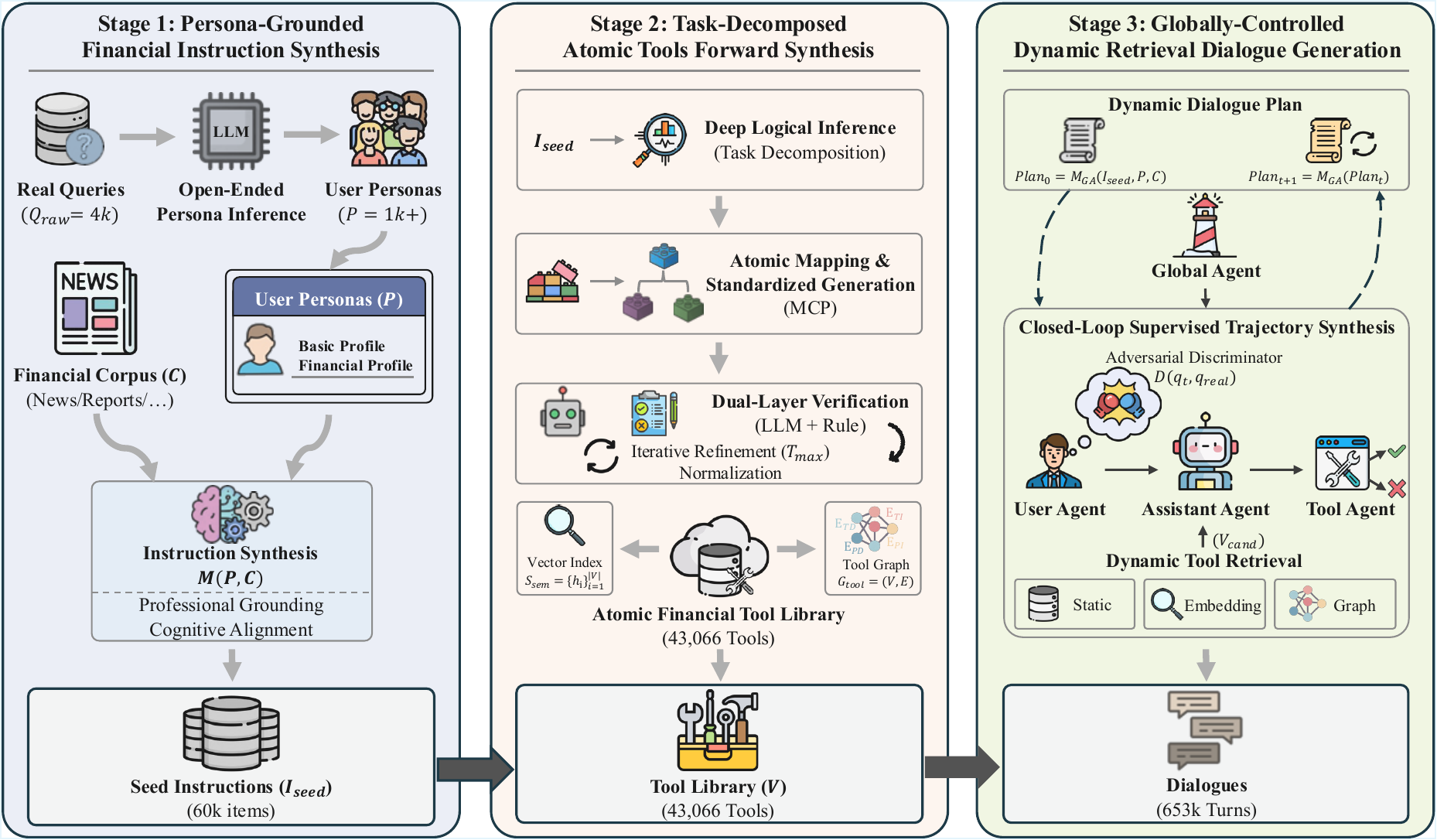}
    \caption{The architecture of FinToolSyn. Our framework employs a forward synthesis paradigm that progresses from persona-based intent generation to dynamic tool retrieval, effectively capturing the implicit and event-driven nature of real-world financial inquiries.}
    \vspace{-4mm}
    \label{fig:framework}
\end{figure*}

\section{FinToolSyn}
\label{sec:FinToolSyn}

As illustrated in Figure \ref{fig:framework}, our approach comprises three core components: 1) persona-grounded financial instruction synthesis to capture diverse user profiles and realistic scenarios; 2) a large-scale, demand-driven atomic tool library designed for fine-grained task decomposition; and 3) a globally-controlled dynamic retrieval mechanism that enables high-fidelity dialogue generation under realistic constraints.

\subsection{Persona-Grounded Financial Instruction Synthesis}
\label{subsec:Persona-Grounded Financial Instruction Synthesis}
Seed instruction quality is the foundation of the forward synthesis pipeline. While we collected 4,000 raw financial queries ($Q_{raw}$) from a large-scale Q\&A platform, these platform-sourced queries often exhibit stylistic homogeneity and platform-specific biases. To capture the diverse, open-ended nature of real-world intent, we introduce a persona-grounded augmentation strategy. By reformulating queries under synthesized User Personas and decoupling intent from product-specific constraints, we ensure the dataset covers a broad spectrum of complex financial goals.

\paragraph{Open-Ended User Persona Inference}
To replicate heterogeneous inquiry patterns, we constructed a latent user space via LLM-based inverse inference on $Q_{raw}$, reconstructing the underlying profile of each authentic request. Each persona $P$ comprises a Basic Profile (demographics and psychological states) and a Financial Profile (e.g., literacy and risk appetite). After semantic de-duplication, we curated 1,000+ unique personas. These explicit profiles constrain the generative model, ensuring synthesized instructions align with assigned expertise and avoiding generic, textbook-style inquiries.

\paragraph{Corpus-Based Financial Instruction Synthesis}
Financial inquiries are inherently event-driven, typically triggered by external information. To ensure substance, we incorporate a financial corpus (e.g., news and earnings reports) as stimulating context $C$. Instruction generation is modeled as $i_{new} \sim P_{\mathcal{M}}(\cdot \mid P, C)$, where LLM $\mathcal{M}$ synthesizes a query by integrating a sampled persona $P$ with context $C$. This mechanism ensures professional grounding by inheriting domain-specific terminology from $C$, and cognitive alignment by calibrating inquiry depth to persona expertise. For example, given the same report, a novice may query general trends while a professional targets metrics like profit margins. This differential synthesis yields a high-fidelity seed set $I_{\text{seed}}$ characterized by expertise-driven nuance.

\subsection{Task-Decomposed Atomic Tools Forward Synthesis}
\label{subsec:Task-Decomposed Atomic Tools Forward Synthesis}
The coverage and granularity of a tool library define the performance ceiling of downstream models. Unlike "supply-driven" datasets that crawl static platforms (e.g., RapidAPI), we adopt a demand-driven principle. By leveraging $I_{\text{seed}}$, we pivot from fixed API catalogs to the latent functional requirements of complex financial tasks. This forward synthesis approach yields an industrial-grade library of atomic tools characterized by professional rigor and fine-grained utility.

\paragraph{Task Decomposition and Tool Synthesis}
To avoid complex "God APIs" that hinder reusability, we implement a three-phase atomic decomposition strategy. First, via Deep Logical Inference, we employ LLMs to derive information interaction chains from $I_{\text{seed}}$. For instance, a "stock valuation" query is decomposed into fetching real-time prices, retrieving fundamental metrics (e.g., EPS and BVPS), and calculating valuation ratios. These composite intents are then decoupled via Atomic Mapping into independent sub-tasks, ensuring each tool adheres to the single-responsibility principle. Finally, Standardized Tool Generation produces documentation adhering to the Model Context Protocol (MCP), enforcing strict type constraints to ensure direct executability.

\paragraph{Dual-Layer Verification and Evolution}
To guarantee robustness, we implement a dual-layer verification pipeline combining LLM-based judgment with rule-based checks. Candidate tools are evaluated across two layers: structural integrity (atomicity and parameter complexity) and logical correctness. Instead of discarding failures, we subject them to iterative refinement guided by diagnostic reports until convergence or reaching $T_{\max}$. Finally, a library normalization phase clusters and merges semantically similar metadata, distilling a unique repository of 43,066 high-quality financial tools.\footnote{See Appendix~\ref{app:api_system} for taxonomy.}

\paragraph{Vector Index and Tool Graph Construction}
To comprehensively simulate the multifaceted retrieval mechanisms in real-world scenarios, we develop two complementary architectures for tool selection. Each tool $v_i \in V$ is characterized by its name $n_i$, description $d_i$, and parameters $\mathcal{P}_i$. \textbf{(1) Semantic Vector Index:} To support global similarity-based search, a pre-trained encoder projects the concatenated text $T_i = \text{Concat}(n_i, d_i)$ into a dense vector ${h}_i$, forming a vector space $S_{\text{sem}}$. This index captures broad query intent and surface-level semantic relevance. \textbf{(2) Tool Dependency Graph:} To capture latent logical associations, we construct a directed graph $G_{\text{tool}} = (V, E)$. To encompass diverse interaction patterns, the edge set $E$ is partitioned into four disjoint subsets: $E = E_{\text{TD}} \cup E_{\text{TI}} \cup E_{\text{PD}} \cup E_{\text{PI}}$. Specifically, Direct Tool Dependency ($E_{\text{TD}}$) defines hard execution prerequisites; Indirect Tool Dependency ($E_{\text{TI}}$) represents soft dependencies for decision optimization; Direct Parameter Dependency ($E_{\text{PD}}$) mandates explicit value extraction; and Indirect Parameter Dependency ($E_{\text{PI}}$) implies parameter determination through logical inference or disambiguation.\footnote{Detailed in Appendix~\ref{app:retrieval_system}.}

\subsection{Globally-Controlled Dynamic Retrieval Dialogue Generation}
\label{subsec:Globally-Controlled Dynamic Retrieval Dialogue Generation}
Existing synthesis paradigms largely rely on static, open-loop generation, in which interaction trajectories are fixed a priori and lack online supervision. Drawing inspiration from control theory, we propose a globally controlled framework that elevates the Global Agent to a dynamic planner and auditor. By continuously validating intermediate user queries and assistant actions against an explicit dialogue plan and evolving context, the framework enables feedback-driven resynthesis during generation. Such closed-loop supervision facilitates adaptive correction and prevents error accumulation in noisy financial environments.

\paragraph{Multi-Agent Simulation Framework}
We employ four specialized agents to construct a high-fidelity interaction ecosystem: (1) the \textbf{User Agent} initiates queries or provides clarifications grounded in assigned personas; (2) the \textbf{Assistant Agent} determines whether to invoke tools from a dynamically retrieved candidate set or respond directly; (3) the \textbf{Tool Agent} simulates execution environments by generating stochastic outputs to reflect real-world API instability; and (4) the \textbf{Global Agent} maintains oversight of the dialogue state and enforces consistency with the dialogue plan. 

\paragraph{Dialogue Plan Generation and Adaptive Updating}
Each synthesized dialogue is initialized with a high-level dialogue plan $\text{Plan}_0$, generated by the Global Agent from three inputs: the synthesis instruction $I_{\text{seed}}$, user persona $P$, and real-time financial context $C$ retrieved via RAG. Rather than serving as a fixed script, the dialogue plan is treated as a mutable state variable $\text{Plan}_t$ that evolves throughout the interaction. As the dialogue progresses, the Global Agent continuously validates intermediate queries and assistant actions against $\text{Plan}_t$ and the current context. When conflicts arise—such as when $\text{Plan}_t$ expects the User Agent to inquire about a price decline but the Assistant Agent reports an increase based on tool outputs—the Global Agent detects the inconsistency and triggers adaptive re-planning, updating $\text{Plan}_t$ to $\text{Plan}_{t+1}$ by revising unexecuted steps to align with the observed evidence. This closed-loop mechanism prevents error accumulation and supports non-linear interaction flows characteristic of realistic financial dialogues.

\begin{table}[!t]
  \centering
  \resizebox{\linewidth}{!}{
  \begin{tabular}{ccccc}
    \toprule
    \textbf{Dataset} & \textbf{\# Tools} & \textbf{\# Dialogues} & \textbf{\# Turns} & \textbf{Avg. Turns} \\
    \midrule
    xLAM~\cite{zhang2024xlamfamilylargeaction} & 3,673& 60,000& 60,000& 1.00\\
    ToolAlpaca~\cite{tang2023toolalpaca} & 426& 3,938& 6,537& 1.66\\
    ToolLLM~\cite{qin2023toolllm} & 16,464& 126,486& 126,486& 1.00\\
    When2Call~\cite{ross2025when2callnottools} & 511& 27,952& 27,952& 1.00\\
    BUTTONInstruct~\cite{BUTTONInstruct} & 13,553& 8,000& 8,000& 1.00\\
    APIGen-MT-5k~\cite{prabhakar2025apigenmtagenticpipelinemultiturn} & 248& 5,000& 24,229& 4.85\\
    ToolACE~\cite{liu2024toolace} & 26,507& 11,300& 27,638& 2.45\\
    ToolMind~\cite{yang2025toolmindtechnicalreportlargescale} & 29,935& 368,611& 794,648& 2.16\\ \midrule
    \textbf{FinToolSyn} & \textbf{43,066} & \textbf{148,984} & \textbf{1,309,630} & \textbf{8.79} \\
    \bottomrule
  \end{tabular}
  }
  \caption{Comparison of FinToolSyn with existing data synthesis resources.}
  \label{tab:constructed_resources}
  \vspace{-4mm}
\end{table}

\paragraph{Closed-Loop Supervised Trajectory Synthesis}
Dialogue synthesis in FinToolSyn follows a closed-loop, supervised process where each agent's output undergoes targeted validation. (1) \textbf{User Agent}: Candidate queries $q_t$ undergo adversarial discrimination via a discriminator $\mathcal{D}$, which evaluates them against real-world samples $q_{\text{real}}$ to filter artificial explicitness, followed by quality inspection and corrective feedback from the Global Agent. (2) \textbf{Assistant Agent}: The Global Agent validates both action decisions (whether to invoke tools from $V_{\text{cand}}$) and action quality (e.g., tool selection and parameter correctness), triggering feedback when violations occur. (3) \textbf{Tool Agent}: Generates stochastic outputs without supervision to preserve real-world API instability. This feedback-driven supervision enables adaptive trajectory correction, preventing error accumulation and distinguishing FinToolSyn from static synthesis paradigms.\footnote{Examples and protocols in Appendix~\ref{app:synthesis_examples}–\ref{app:global_control_examples}.}

\paragraph{Diversified and Noisy Retrieval Environments}
To reflect realistic deployment conditions and evaluate model robustness, we implement three retrieval configurations for $V_{\text{cand}}$: (1) \textbf{Static Retrieval} uses a fixed candidate set pre-determined by the Global Agent based on $\text{Plan}_0$, simulating stable closed-domain scenarios. (2) \textbf{Vector Retrieval} employs turn-level semantic search with query rewriting to resolve coreferences in multi-turn contexts. (3) \textbf{Graph-Enhanced Retrieval} augments vector search with $K$-hop neighbors from the tool dependency graph $G_{\text{tool}}$, introducing hard negatives that compel precise discrimination amidst dense functional noise. Candidate sets may include irrelevant, partially relevant, or redundant tools, and their composition varies dynamically across turns. Coupled with closed-loop supervision, this noisy environment naturally induces exploration, recovery, and revision behaviors that approximate real-world financial decision-making.
Throughout the entire synthesis pipeline, a strict quality control protocol is enforced: the overall discard rate is kept below 10\%, and a sampled subset of the synthesized data achieves a human acceptance rate of 94.2\% from finance-specialized annotators. Full details on filtering criteria, failure cases, and human verification are provided in Appendix~\ref{app:data_quality}.

\begin{table}[!t]
  \centering
  \resizebox{\columnwidth}{!}{
  \begin{tabular}{l l l}
      \toprule
      \textbf{Class} & \textbf{Tool Setting} & \textbf{Response Type} \\
      \midrule
      \multirow{2}{*}{\textbf{Tool-Call}} 
          & Single-tool & Single, Parallel \\
          \cmidrule(l){2-3}
          & Multi-tool & Single, Parallel, Serial \\
      \midrule
      \textbf{Non-Tool-Call} 
          & Multi-tool & UD, CI, DR \\
      \bottomrule
  \end{tabular}
  }
  \caption{Hierarchical classification of FinToolBench. UD: Unavailability Detection; CI: Clarification Inquiry; DR: Direct Response.}
  \label{tab:taxonomy_compact}
  \vspace{-4mm}
\end{table}

\begin{table*}[t]
\centering
\renewcommand{\arraystretch}{0.85} %
\resizebox{\linewidth}{!}{
\scriptsize %
\begin{tabular}{lccccccccccc} 
\toprule
\multirow{2}{*}{\textbf{Models}} & \multirow{2}{*}{\textbf{Avg}} & \multicolumn{4}{c}{\textbf{Tool-Call}} & \multicolumn{3}{c}{\textbf{Non-Tool-Call}} & \multicolumn{2}{c}{\textbf{Financial Indicators}} \\ \cmidrule(lr){3-6} \cmidrule(lr){7-9} \cmidrule(lr){10-11}
 & & \textbf{ST-SC} & \textbf{ST-MC} & \textbf{MT-SC} & \textbf{MT-MC} & \textbf{UD} & \textbf{CI} & \textbf{DR} & \textbf{ITA} & \textbf{KDA} \\ 
\midrule

\multicolumn{11}{c}{\cellcolor{gray!10}\textbf{Panel A: Prompt Mode}} \\ 
\multicolumn{11}{r}{\textit{Closed-Source LLMs}} \\
Gemini-2.5-Flash & 40.85 & 1.92 & 4.60 & 22.22 & 18.71 & \best{100.00} & 38.46 & \best{100.00} & 25.93 & 10.25 \\
Gemini-2.5-Pro & \best{62.15} & \best{81.56} & \best{72.34} & 67.32 & 67.27 & 54.64 & 25.00 & 66.90 & 14.81 & \best{72.75} \\
Claude-4.0-Sonnet & 57.96 & 58.09 & 37.59 & 37.09 & 43.98 & \best{100.00} & 31.73 & \secbest{97.25} & 37.04 & 44.03 \\
GPT-3.5-Turbo & 38.47 & 56.77 & 39.48 & 42.75 & 37.26 & 41.72 & 4.81 & 46.50 & 11.11 & 45.04 \\
GPT-4o & \secbest{59.67} & 74.87 & 58.02 & 62.36 & 53.98 & 65.47 & 28.85 & 74.18 & 37.04 & 61.99 \\ 
\multicolumn{11}{r}{\textit{Open-Source LLMs}} \\
Deepseek-V3.1-Terminus & 55.32 & \secbest{79.83} & 56.83 & 73.61 & 62.82 & 47.81 & 3.85 & 62.50 & 3.70 & 67.04 \\
Qwen3-235B-Instruct & 53.05 & 77.19 & 59.11 & 67.79 & 66.50 & 43.58 & 1.92 & 55.22 & 0.00 & 67.02 \\
Qwen2.5-72B-Instruct & 51.86 & 78.12 & \secbest{59.69} & 64.79 & 62.84 & 36.15 & 5.77 & 55.63 & 11.11 & 66.92 \\
Qwen2.5-32B-Instruct & 51.18 & 66.26 & 45.39 & 53.43 & 51.33 & 65.08 & 3.85 & 72.94 & 3.70 & 53.81 \\
Qwen2.5-14B-Instruct & 55.75 & 68.47 & 52.06 & 54.91 & 51.93 & 76.74 & 6.73 & 79.40 & 3.70 & 56.99 \\
Qwen2.5-7B-Instruct & 41.67 & 61.10 & 45.12 & 60.17 & 47.93 & 30.81 & 0.00 & 46.57 & 0.00 & 51.67 \\
Qwen3-32B & 51.67 & 79.82 & 57.85 & 63.60 & 64.79 & 36.92 & 3.85 & 54.88 & 0.00 & 66.20 \\
Qwen3-14B & 47.87 & 67.43 & 51.69 & 64.68 & 58.58 & 36.21 & 1.92 & 54.60 & 0.00 & 58.59 \\
Qwen3-8B & 47.55 & 70.24 & 47.04 & 59.42 & 53.71 & 46.09 & 1.92 & 54.46 & 3.70 & 57.37 \\ 
\multicolumn{11}{r}{\textit{FinTool-Tuned (Ours)}} \\
FinTool-Qwen3-14B & 57.95 & 77.23 & 58.24 & \best{79.99} & \best{68.78} & 30.32 & \best{50.00} & 41.07 & \best{74.07} & \secbest{68.82} \\
FinTool-Qwen3-8B & 55.68 & 78.10 & 58.12 & \secbest{79.37} & \secbest{67.82} & 30.32 & \secbest{42.31} & 33.72 & \secbest{70.37} & 67.79 \\ 

\midrule

\multicolumn{11}{c}{\cellcolor{gray!10}\textbf{Panel B: Function Calling (FC) Mode}} \\ 
\multicolumn{11}{r}{\textit{Closed-Source LLMs}} \\
Gemini-2.5-Flash & \secbest{58.35} & 57.73 & 28.50 & 42.80 & 38.80 & \best{94.27} & \best{50.37} & \best{96.00} & 43.48 & 41.66 \\
Gemini-2.5-Pro & \best{60.98} & 64.77 & 47.68 & 49.56 & 51.17 & \secbest{85.07} & 40.75 & \secbest{87.83} & 34.78 & 53.94 \\
Claude-4.0-Sonnet & 55.93 & 67.28 & \best{62.15} & 68.27 & 63.36 & 44.70 & 32.65 & 53.08 & 44.44 & 65.33 \\
GPT-3.5-Turbo & 43.67 & 68.51 & 37.15 & 53.36 & 45.78 & 44.78 & 4.17 & 51.91 & 0.00 & 48.94 \\
GPT-4o & 58.26 & 74.53 & 53.25 & 71.56 & 56.10 & 63.01 & 6.25 & 83.13 & 11.11 & 62.59 \\ 
\multicolumn{11}{r}{\textit{Open-Source LLMs}} \\
Qwen3-235B-Instruct & 58.06 & 76.98 & 59.83 & 64.04 & \best{67.14} & 59.47 & 9.62 & 69.37 & 3.70 & 66.61 \\
Qwen2.5-72B-Instruct & 53.23 & \best{79.12} & \secbest{61.23} & 66.45 & 64.12 & 38.23 & 6.12 & 57.34 & 11.50 & 67.10 \\
Qwen2.5-32B-Instruct & 52.85 & 68.23 & 47.12 & 55.45 & 53.12 & 67.34 & 4.12 & 74.56 & 4.10 & 54.20 \\
Qwen2.5-14B-Instruct & 57.23 & 70.12 & 54.23 & 56.45 & 53.12 & 78.34 & 7.12 & 81.23 & 4.10 & 57.50 \\
Qwen2.5-7B-Instruct & 43.23 & 63.12 & 46.23 & 62.45 & 49.12 & 32.34 & 1.12 & 48.23 & 1.10 & 52.10 \\
Qwen3-32B & 49.58 & 73.13 & 50.14 & 48.23 & 51.49 & 53.09 & 3.85 & 67.10 & 0.00 & 54.71 \\
Qwen3-14B & 50.47 & 68.25 & 51.13 & 43.64 & 44.39 & 63.75 & 4.81 & 77.34 & 0.00 & 53.13 \\
Qwen3-8B & 50.05 & 66.55 & 43.37 & 39.22 & 42.97 & 73.42 & 3.85 & 80.98 & 3.70 & 49.05 \\ 
\multicolumn{11}{r}{\textit{FinTool-Tuned (Ours)}} \\
FinTool-Qwen3-14B & 57.77 & \secbest{77.86} & 59.11 & \best{76.98} & \secbest{66.18} & 33.32 & \secbest{45.19} & 45.74 & \best{74.07} & \best{68.17} \\
FinTool-Qwen3-8B & 57.64 & 77.20 & 56.01 & 74.53 & 65.50 & 42.98 & 44.23 & 42.99 & \secbest{59.26} & 66.26 \\

\bottomrule
\end{tabular}
}
\caption{Main results of FinToolBench. The table is split into \textbf{Panel A} (Prompt Mode) and \textbf{Panel B} (Function Calling Mode). The \best{best} and \secbest{second-best} results are highlighted separately for each panel.}
\label{tab:main_results}
\vspace{-4mm}
\end{table*}

\section{Experimental Setup}

\subsection{Constructed Resources}

FinToolSyn produces 148,984 dialogues over 1.3M turns from 43,066 instruments, as detailed in Table~\ref{tab:constructed_resources}.\footnote{Extended statistics in Appendix~\ref{app:dataset_distribution}.} The synthesis is seeded with 1,000+ user profiles and 60,000 instructions, where multi-agent simulation captures realistic, multi-turn financial interactions.

\subsection{Financial Tool-Use Benchmark}
\label{Financial Tool-Use Benchmark}
We introduce \textbf{FinToolBench}, comprising 843 gold-standard instances verified by finance-specialized annotators.\footnote{Guidelines in Appendix~\ref{app:Annotation Guidelines for FinToolBench Verification}.} The benchmark evaluates model performance across two orthogonal dimensions and multiple response categories, as illustrated in Table~\ref{tab:taxonomy_compact}.

\paragraph{Benchmark Configurations}
FinToolBench evaluation is structured along two independent dimensions: \textbf{Context Depth}—Single-Turn (ST) versus Multi-Turn (MT) interactions; and \textbf{Candidate Tool Density}—Single-Candidate (SC) versus Multi-Candidate (MC) retrieval scenarios.

Tool-Call instances are evaluated across four configurations formed by the Cartesian product of these dimensions: ST-SC, ST-MC, MT-SC, and MT-MC. Non-Tool-Call instances assess the model's decision logic through three response categories: \textbf{Unavailability Detection (UD)} evaluates whether models correctly identify scenarios lacking suitable tools; \textbf{Clarification Inquiry (CI)} measures the ability to explicitly request missing parameters; and \textbf{Direct Response (DR)} assesses accuracy in generating conclusive answers when sufficient context is already available.

All configurations are evaluated under two paradigms: \textbf{Prompt Mode}, where tool schemas are serialized as a text block injected via a system prompt (Figure~\ref{fig:system_prompt_for_prompt_mode}), and \textbf{Function Calling (FC) Mode}, where schemas are passed as native JSON metadata via the model's built-in API (Figure~\ref{fig:mcp_to_fc}).

\paragraph{Comparison with Existing Benchmarks}
Table~\ref{tab:benchmark_comparison} positions FinToolBench against representative general-purpose benchmarks. Existing benchmarks such as BFCL~\cite{BFCL} predominantly evaluate models using oracle toolsets (providing exactly the needed tools), while $\tau$-bench~\cite{yao2024taubenchbenchmarktoolagentuserinteraction} restricts agents to fewer than 20 APIs per domain. FinToolBench uniquely combines a massive dynamic retrieval environment with domain-specific evaluation protocols, enabling rigorous assessment of the capabilities that matter most in real-world financial deployment.

\begin{table}[!t]
\centering
\small
\setlength{\tabcolsep}{3pt}
\resizebox{\linewidth}{!}{
\begin{tabular}{lccc}
\toprule
\textbf{Dimension} & \textbf{BFCL} & \textbf{$\tau$-bench} & \textbf{FinToolBench} \\
\midrule
Tool Space & Small \& Oracle & $<$20 APIs/domain & 43,066 domain tools \\
Data Style & Explicit & Task-driven & Underspecified, event-driven \\
Candidate Env. & Static \& given & Fixed domain & Dynamic + hard negatives \\
Task Focus & Syntax \& exec. & State tracking & Intent elicitation \& navigation \\
Finance-Specific & \texttimes & \texttimes & \checkmark \\
Key Metrics & AST Match & Task Success & CB-HWS, KDA, ITA \\
\bottomrule
\end{tabular}
}
\caption{Comparison of FinToolBench with existing benchmarks.}
\label{tab:benchmark_comparison}
\vspace{-6mm}
\end{table}

\paragraph{Circuit-Breaker Hierarchical Weighted Scoring}
\label{CBHWS}
We propose the Circuit-Breaker Hierarchical Weighted Scoring (CB-HWS) to quantify model reliability in high-stakes financial settings. The mechanism cascades from hard rule-based constraints to soft LLM-based evaluation, implementing a ``fail-fast'' principle where any critical violation immediately terminates scoring. Let $\mathcal{T}_{\text{gold}}$ and $\mathcal{T}_{\text{pred}}$ denote ground truth and predicted action sequences; the total score $S_{\text{total}} \in [0, 100]$ branches by instance type:
\begin{equation}
S_{\text{total}} =
\begin{cases}
S_{\text{TC}}(\mathcal{T}_{\text{pred}}, \mathcal{T}_{\text{gold}}) & \text{if Tool-Call} \\
S_{\text{NTC}}(\mathcal{T}_{\text{pred}}) & \text{if Non-Tool-Call}
\end{cases}
\end{equation}

\textbf{\emph{Tool-Call Scoring ($S_{\text{TC}}$).}}
Scoring proceeds in two sequential phases.

\textbf{Phase 1 --- Rule-Based Circuit Breaker.} Three hard checks are applied in order; failure at any step immediately yields a score of 0:
(1)~\textit{Format Compliance}: the prediction must contain at least one tool call in the required schema format;
(2)~\textit{Tool Hallucination Check}: every invoked tool name must exist in the candidate set $V_{\text{cand}}$;
(3)~\textit{Parameter Schema Compliance}: all required parameters must be present, parameter names must match the schema, and value types must satisfy defined constraints (e.g., \texttt{array}, \texttt{enum}). Only when all three checks pass does the rule indicator $V_{\text{rule}}=1$ and scoring proceeds.

\textbf{Phase 2 --- LLM-Based Soft Evaluation.} An LLM judge evaluates execution quality in two steps. First, it assigns a Tool Selection Score $k \in [0, 10]$ measuring whether the invoked tool combination fully covers the user intent without redundancy. If $k = 0$, a logical circuit break is triggered and the score is 0. Otherwise, for each invoked tool $i$, the judge produces a Parameter Name Score $x_i \in [0, 10]$ (assessing whether the chosen parameter set is complete and appropriate) and per-parameter value scores $y_{ij} \in [0, 10]$, with $\bar{y}_i = \frac{1}{m_i}\sum_{j=1}^{m_i} y_{ij}$.

\textbf{Hierarchical Aggregation.} The per-tool execution score is:
\vspace{-4mm}

\begin{equation}
s_i = \begin{cases} w_{\text{struct}} \cdot x_i + w_{\text{val}} \cdot \bar{y}_i & \text{if } m_i > 0 \\ x_i & \text{if } m_i = 0 \end{cases}
\end{equation}
The overall execution score, turn score, and final multi-turn aggregate are:
\vspace{-5mm}

\begin{gather}
S_{\text{exec}} = \frac{1}{N}\sum_{i=1}^{N} s_i \\[4pt]
S_{\text{turn}} = \left[w_{\text{select}} \cdot k + w_{\text{exec}} \cdot S_{\text{exec}}\right] \times 10 \\[4pt]
S_{\text{TC}} = \frac{1}{T}\sum_{t=1}^{T} S_{\text{turn}}^{(t)}
\label{equation:V}
\end{gather}
where $T$ is the number of interaction turns (for single/parallel calls, $T{=}1$; for serial calls, $T{>}1$). We set $w_{\text{val}}=0.7$, $w_{\text{struct}}=0.3$, $w_{\text{exec}}=0.6$, $w_{\text{select}}=0.4$, reflecting the primacy of value accuracy in financial contexts. Complete pseudocode is provided in Appendix~\ref{app:cbhws_algo}.

\textbf{\emph{Non-Tool-Call Scoring ($S_{\text{NTC}}$).}}
For instances prohibiting tool invocation, scoring enforces strict rejection and intent recognition. Any hallucinated tool call immediately yields a score of 0. For predictions without tool calls:
\begin{equation}
S_{\text{NTC}} = \mathbb{I}(N_{\text{pred}}=0) \cdot \delta_{\text{intent}} \cdot 100
\end{equation}
where $N_{\text{pred}}$ is the predicted tool call count and:
\begin{equation}
\resizebox{0.99\columnwidth}{!}{%
$\delta_{\text{intent}} =
\begin{cases}
1 & \text{for UD and DR instances} \\
\mathbb{I}(\text{Judge confirms clarification intent}) & \text{for CI instances}
\end{cases}$
}
\end{equation}
For UD and DR, the absence of tool calls suffices for a full score. For CI, the model must explicitly request the missing parameters necessary for tool execution, as verified by an LLM Judge; otherwise the score is 0.

\subsection{Evaluation Metrics}
\label{Evaluation Metrics}
For FinToolBench, we report the comprehensive CB-HWS score alongside two domain-specific metrics. \textbf{Key Digit Accuracy (KDA)} measures exact matches for critical values (e.g., transaction amounts) where minor deviations cause substantial errors. \textbf{Invocation Timing Accuracy (ITA)} assesses adherence to logical business workflows (e.g., risk verification). These metrics provide complementary domain-specific dimensions.

To evaluate broader capabilities, we assess two aspects: (1) \textbf{Transferability to General Tool Use}, measured by the BFCL-v4 suite and $\tau$-bench (Pass@1); and (2) \textbf{Preservation of General Reasoning}, monitored via GSM8K (math), MMLU (knowledge), and HumanEval (coding) to ensure domain adaptation does not compromise intrinsic abilities.

\subsection{Implementation Details}
\paragraph{Data Synthesis}
The pipeline used DeepSeek-V3.1-Terminus as the global agent orchestrating the synthesis workflow, with Qwen3-235B-A22B-Instruct-2507 instantiating User, Assistant, and Tool agents (temperature 0.6). The retriever recalled top-10 candidates ($K=10$) per query, with $T_{\max}=3$ retries for self-correction.

\paragraph{Model Training}
LoRA-based SFT on four LLMs ($<$14B parameters) using 100k sampled instances. Training: 8 H20 GPUs, 1 epoch, 32k max sequence length, batch size 128, learning rate 2e-4, warmup 0.1. Inference used greedy decoding with official chat templates.

\paragraph{Evaluation}
All inference used temperature 0 and 9,182 max tokens for deterministic results. DeepSeek-V3.2 served as the Judge model for CB-HWS evaluation, assessing tool selection accuracy, parameter correctness, and CI intent verification.\footnote{Evaluation prompts in Appendix~\ref{app:Evaluation Prompt Design}.}

\section{Experiments and Analysis}

\subsection{Main Results}
Table~\ref{tab:main_results} presents the performance on FinToolBench, organized into Panel A (Prompt Mode) and Panel B (Function Calling Mode). The results demonstrate that our framework effectively bridges the gap between open-source models and frontier proprietary systems.\footnote{Full results in Appendix~\ref{app:total_results}.}

\paragraph{Robustness in Complex Multi-Turn Scenarios.} Our fine-tuned models exhibit exceptional stability in long-horizon interactions. In Panel A, FinTool-Qwen3-14B surpasses GPT-4o in Multi-Turn (e.g., 79.99 in MT-SC). Similarly, in Panel B, the FC-tuned version maintains this dominance (76.98), whereas base models like Qwen3-8B degrade significantly (dropping to 39.22). This confirms that the closed-loop trajectory synthesis in FinToolSyn effectively mitigates error accumulation in complex financial contexts.

\paragraph{Mastery of Financial Business Logic.} A critical disparity exists in the Financial Indicators. Base models frequently score 0.00 in Invocation Timing Accuracy (ITA), indicating a failure to verify prerequisites (e.g., risk checks). In contrast, our models achieve up to 74.07, significantly outperforming Claude-4.0-Sonnet (44.44). Consistently high Key Digit Accuracy (KDA) further proves that our models prioritize logical correctness and numeric precision over simple API formatting.\footnote{Qualitative case study in Appendix~\ref{app:case_study}.}

\paragraph{Versatility Across Interaction Paradigms.} FinToolSyn proves effective in synthesizing training corpora for distinct paradigms. Models trained on our Prompt-formatted data (Panel A) show substantial gains over base versions. Crucially, our FC-trained models (Panel B) adapt to strict structural constraints where general models often falter (e.g., Qwen3-32B drops performance in FC mode), achieving stability and accuracy comparable to proprietary models like GPT-4o.

\subsection{Fine-Grained Error Analysis}
\label{subsec:error_analysis}

To provide deeper insight into failure modes, we decompose prediction errors across multiple dimensions and compare FinTool-Qwen3-14B against baselines under both FC and Prompt modes in Table~\ref{tab:error_analysis}.

\begin{table}[!t]
\centering
\small
\setlength{\tabcolsep}{3pt}
\resizebox{\linewidth}{!}{
\begin{tabular}{lcccccc}
\toprule
\multirow{2}{*}{\textbf{Metric}} & \multicolumn{3}{c}{\textbf{FC Mode}} & \multicolumn{3}{c}{\textbf{Prompt Mode}} \\
\cmidrule(lr){2-4}\cmidrule(lr){5-7}
 & \textbf{GPT-4o} & \textbf{Qwen3-14B} & \textbf{FinTool-14B} & \textbf{GPT-4o} & \textbf{Qwen3-14B} & \textbf{FinTool-14B} \\
\midrule
\multicolumn{7}{l}{\textit{Non-Tool Scenarios}} \\
Total Error Rate       & 38.2\% & 51.4\% & 58.7\% & 43.8\% & 68.9\% & 59.7\% \\
~~Invalid Invocation   & 34.7\% & 41.9\% & 56.8\% & 35.2\% & 64.8\% & 59.0\% \\
~~Intent/CI Failure    & 3.5\%  & 9.5\%  & \textbf{1.9\%}  & 8.6\%  & 4.1\%  & \textbf{0.6\%}  \\
Correct Rate           & \textbf{61.8\%} & 48.6\% & 41.3\% & \textbf{56.2\%} & 31.1\% & 40.3\% \\
\midrule
\multicolumn{7}{l}{\textit{Tool-Call Scenarios}} \\
Rule Check Failed      & 12.4\% & 28.0\% & \textbf{3.6\%}  & 19.3\% & 15.3\% & \textbf{3.6\%}  \\
~~Format Error         & 9.9\%  & 23.9\% & \textbf{1.7\%}  & 16.9\% & 9.1\%  & \textbf{1.5\%}  \\
~~Tool Hallucination   & 0.0\%  & 0.0\%  & \textbf{0.0\%}  & 0.4\%  & 1.9\%  & \textbf{0.0\%}  \\
~~Parameter Error      & 2.5\%  & 4.2\%  & \textbf{1.9\%}  & \textbf{2.1\%} & 4.4\%  & \textbf{2.1\%}  \\
Rule Check Passed      & 87.6\% & 72.0\% & \textbf{96.4\%} & 80.7\% & 84.7\% & \textbf{96.4\%} \\
~~Tool Selection (Max=10)     & 7.36 & 7.67 & \textbf{7.81} & 7.44 & \textbf{7.93} & 7.65 \\
~~Param.\ Structure (Max=10)  & 9.48 & \textbf{9.52} & 9.50 & 9.60 & \textbf{9.65} & 9.51 \\
~~Param.\ Value (Max=10)      & \textbf{8.49} & 8.16 & 8.41 & \textbf{8.67} & 8.12 & 8.36 \\
\bottomrule
\end{tabular}
}
\caption{Fine-grained error distribution comparing FinTool-Qwen3-14B against baselines. \textbf{Bold} indicates best performance per column group.}
\label{tab:error_analysis}
\vspace{-4mm}
\end{table}

\paragraph{Structural Reliability.} FinTool-Qwen3-14B achieves the lowest Rule Check Failed rate (3.6\%) across both modes, particularly excelling in format compliance (Format Error: 1.7\% in FC mode vs.\ 9.9\% for GPT-4o). The zero tool hallucination rate confirms that domain-specific fine-tuning instills a reliable grounding in the candidate tool space.

\paragraph{Intent Discrimination Trade-off.} A notable trade-off emerges in Non-Tool scenarios: FinTool exhibits a higher overall error rate (58.7\%) driven by over-triggering (Invalid Invocation: 56.8\%), a ``pro-tool bias'' induced by training on a corpus rich in tool-use dialogues. However, FinTool achieves the lowest Intent/Clarification Failure rate (1.9\% FC, 0.6\% Prompt), demonstrating superior understanding of \textit{when} clarification is needed. This reveals a nuanced failure mode: the model correctly recognizes the need for more information but sometimes still attempts a speculative tool call rather than asking. Mitigating this over-triggering behavior is a direction for future alignment work.

\subsection{Generalization Capabilities}

\begin{figure}[!t]
    \centering
    \resizebox{0.89\linewidth}{!}{\includegraphics{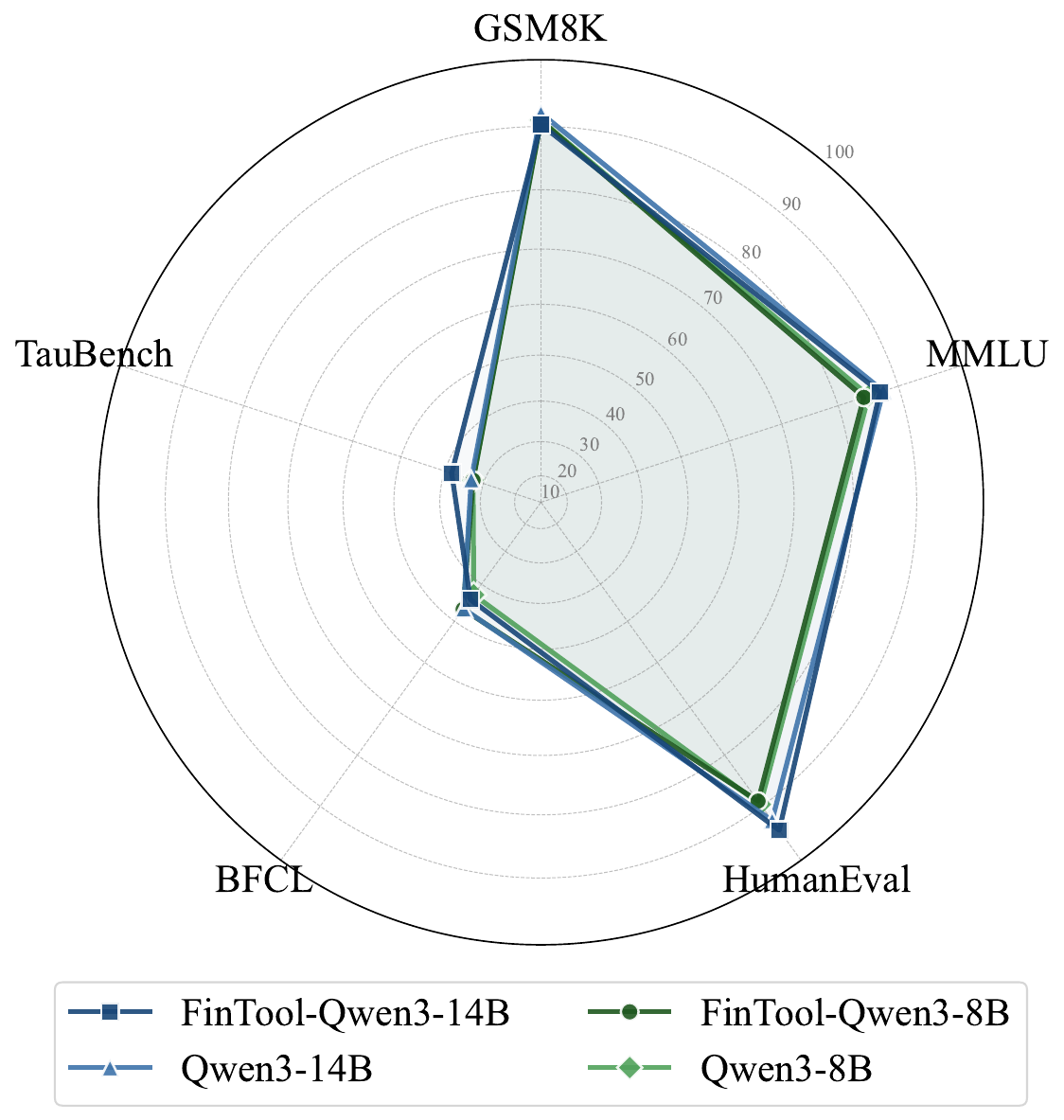}}
    \caption{Performance comparison of various general capabilities before and after training different models.}
    \vspace{-4mm}
    \label{fig:Generalization_Capabilities}
\end{figure}

To verify that FinToolSyn instills robust reasoning patterns rather than mere dataset memorization, we evaluated the models on general-purpose benchmarks (Figure~\ref{fig:Generalization_Capabilities}). Detailed numeric results with per-model deltas are provided in Table~\ref{tab:generalization_numeric}.

\paragraph{Transferability to General Tool Use.} On the BFCL and $\tau$-bench benchmarks, our fine-tuned models perform on par with or slightly surpass their base counterparts. FinTool-Qwen3-8B improves BFCL by +3.9 points and $\tau$-bench by 0 (maintaining parity), while FinTool-Qwen3-14B achieves a +5.0 point gain on $\tau$-bench. This indicates that the model has internalized the abstract structure of tool invocation—such as parameter extraction and intent mapping—allowing it to transfer these skills to non-financial domains without overfitting to specific API schemas.

\paragraph{Preservation of General Reasoning.} Domain-specific fine-tuning often risks catastrophic forgetting. However, on GSM8K, HumanEval, and MMLU, our models exhibit negligible performance fluctuation (all within $\pm$2 points). Notably, FinTool-Qwen3-14B even improves HumanEval by +1.83 points. This stability suggests that FinToolSyn’s high-quality, logically coherent synthetic data enhances domain competence while preserving the model’s reasoning backbone.

\subsection{Ablation Studies}
\begin{table}[!t]
\centering
\small
\setlength{\tabcolsep}{3.5pt}
\resizebox{\linewidth}{!}{
\begin{tabular}{lccccc}
\toprule
\multirow{2}{*}{\textbf{Model Setting}} & \multicolumn{2}{c}{\textbf{Methodology}} & \multicolumn{3}{c}{\textbf{Performance}} \\
\cmidrule(lr){2-3} \cmidrule(lr){4-6}
& \textbf{Synthesis} & \textbf{Retrieval} & \textbf{Tool-Call} & \textbf{Non-Tool} & \textbf{Avg.} \\
\midrule
Base (Qwen3-8B) & - & - & 48.05 & \textbf{52.75} & 50.40 \\
\midrule
Reverse Syn. & Reverse & Static & 64.01 & 15.01 & 39.51 \\
FinToolSyn (Static) & \textbf{Forward} & Static & 65.84 & 44.03 & 54.93 \\
\rowcolor{gray!10} \textbf{FinToolSyn (Full)} & \textbf{Forward} & \textbf{Dynamic} & \textbf{68.31} & 43.40 & \textbf{55.86} \\
\bottomrule
\end{tabular}
}
\caption{\textbf{Ablation study.} Reverse Synthesis improves tool calling but degrades general logic (Non-Tool score drops). FinToolSyn restores logic and enhances robustness via dynamic retrieval.}

\label{tab:ablation}
\vspace{-4mm}
\end{table}

\paragraph{Synthesis Directionality.} Table~\ref{tab:ablation} shows while Reverse Synthesis improves tool usage, it introduces ``artificial explicitness'' bias, crashing Non-Tool accuracy to 15.01. Conversely, Forward Synthesis aligns user needs with solutions, restoring decision logic (Non-Tool: 44.03) and boosting Tool-Call performance (65.84). This proves our method teaches \textit{when} to use tools, not just \textit{how}.

\paragraph{Retrieval Strategy.}
Dynamic Retrieval enhances robustness against noise. By training with evolving, imperfect candidate sets, the model learns to filter distractors rather than passively execute instructions. This mechanism pushes the Tool-Call score to 68.31 and achieves the best overall performance (55.86), confirming that simulating retrieval noise is crucial for agent resilience.

\section{Conclusion}
In this paper, we introduce FinToolSyn to synthesize high-quality data for financial agents. By combining Forward Synthesis and Dynamic Retrieval, it mitigates the ``artificial explicitness'' bias of backward methods. Experiments demonstrate that FinToolSyn outperforms baselines, achieving a critical balance between mastering complex tools and retaining the reasoning logic to reject invalid requests, enabling more reliable agents.

\section*{Limitations}
Despite the robustness and high fidelity of our framework, its reliance on domain-specific tool taxonomies and dependency graphs means that adapting the system to other specialized fields, such as medicine or law, would require non-trivial expert-driven re-calibration. Furthermore, the sophisticated multi-agent orchestration and iterative adversarial filtering loops, while essential for mitigating artificial explicitness, inevitably incur higher computational overhead during the data synthesis phase compared to conventional open-loop generation. Finally, although our environment effectively simulates plan-fact conflicts and stochastic tool failures, a subtle gap remains between our synthetic simulation and the extreme corner cases or real-time latency fluctuations inherent in production-grade financial APIs.

\bibliography{custom}

\clearpage 

\appendix
\section{Data Synthesis Details}
\label{app:data_synthesis}

\subsection{API Relationship Extraction System Documentation}
\label{app:retrieval_system}
This appendix provides detailed documentation for the API relationship extraction system described in Section~\ref{subsec:Task-Decomposed Atomic Tools Forward Synthesis}. We define four types of dependencies and illustrate each with concrete examples from financial domain APIs.

\subsubsection{Dependency Type Definitions}

\paragraph{Direct Parameter Dependency }

A direct parameter dependency exists when an input parameter of one tool must be provided by the output of another tool, and this parameter cannot be directly supplied by the user. This occurs when API $A$'s output field exactly matches API $B$'s required input parameter, where the parameter represents a system-generated internal identifier.

\begin{toolbox}[title={\textbf{API A: Company Name Fuzzy Search}}]
\begin{lstlisting}[style=jsonstyle]
{
    "name": "search_company_by_name",
    "description": "Fuzzy search by company name keywords, returns matching company codes (company_code)",
    "inputSchema": {
        "properties": {
            "keyword": {
                "type": "string", 
                "description": "Company name keyword"
            }
        },
        "required": ["keyword"]
    },
    "outputSchema": {
        "properties": {
            "company_code": {
                "type": "string", 
                "description": "Internal system company code"
            },
            "company_full_name": {"type": "string"}
        }
    }
}
\end{lstlisting}
\end{toolbox}

\begin{toolbox}[title={\textbf{API B: Company Registration Information Query}}]
\begin{lstlisting}[style=jsonstyle]
{
    "name": "get_company_registration_info",
    "description": "Query company registration information by company code, including registered capital, legal representative, business scope, etc.",
    "inputSchema": {
        "properties": {
            "company_code": {
                "type": "string", 
                "description": "Internal system company code"
            }
        },
        "required": ["company_code"]
    }
}
\end{lstlisting}
\end{toolbox}

\begin{toolbox}[title={\textbf{Extracted Relationship}}]
\begin{lstlisting}[style=jsonstyle]
{
    "head": "search_company_by_name",
    "relation": "direct_parameter_dependency",
    "tail": "get_company_registration_info"
}
\end{lstlisting}
\end{toolbox}

\noindent\textbf{Explanation:} The API \texttt{get\_company\allowbreak\_registration\_info} requires \texttt{company\_code} (internal system company code) as a mandatory parameter. This is a system-generated internal identifier that users cannot know in advance, thus requiring a prior call to \texttt{search\_company\_by\_name} to obtain it through company name search.

\paragraph{Indirect Parameter Dependency}

An indirect parameter dependency exists when an input parameter of one tool may benefit from another tool for resolution or enrichment in specific contexts, but is not mandatory. The input parameter can be directly provided by the user, but a more accurate or standardized value can be obtained through another API.

\begin{toolbox}[title={\textbf{API A: Stock Code Search}}]
\begin{lstlisting}[style=jsonstyle]
{
    "name": "search_stock_code",
    "description": "Query A-share stock code by company abbreviation or keyword",
    "inputSchema": {
        "properties": {
            "keyword": {"type": "string"}
        },
        "required": ["keyword"]
    },
    "outputSchema": {
        "properties": {
            "stock_code": {
                "type": "string", 
                "description": "6-digit stock code"
            },
            "stock_name": {"type": "string"},
            "exchange": {
                "type": "string", 
                "description": "Exchange: SSE or SZSE"
            }
        }
    }
}
\end{lstlisting}
\end{toolbox}

\begin{toolbox}[title={\textbf{API B: Stock Financial Metrics Query}}]
\begin{lstlisting}[style=jsonstyle]
{
    "name": "get_stock_financial_metrics",
    "description": "Query core financial metrics of a stock, including P/E ratio, P/B ratio, ROE, etc. Supports query by stock code or name",
    "inputSchema": {
        "properties": {
            "stock_code": {
                "type": "string", 
                "description": "6-digit stock code, e.g., 600519"
            },
            "stock_name": {
                "type": "string", 
                "description": "Stock abbreviation, e.g., Kweichow Moutai"
            }
        },
        "required": []
    }
}
\end{lstlisting}
\end{toolbox}

\begin{toolbox}[title={\textbf{Extracted Relationship}}]
\begin{lstlisting}[style=jsonstyle]
{
    "head": "search_stock_code",
    "relation": "indirect_parameter_dependency",
    "tail": "get_stock_financial_metrics"
}
\end{lstlisting}
\end{toolbox}

\noindent\textbf{Explanation:} The API \texttt{get\_stock\allowbreak\_financial\_metrics} can query directly using stock abbreviations like ``Kweichow Moutai,'' which users typically know. However, first obtaining the accurate stock code ``600519'' through \texttt{search\_stock\_code} can avoid query errors caused by abbreviation ambiguity (e.g., ``Ping An of China'' vs. ``Ping An Bank''). This represents a ``nice-to-have'' enhancement relationship.

\paragraph{Direct Tool Dependency}

A direct tool dependency exists when a tool's execution strictly requires another tool as a prerequisite, where the description explicitly states that a prior operation must be performed, or that an analysis must be based on another analysis's results.

\begin{toolbox}[title={\textbf{API A: Get Stock K-line History}}]
\begin{lstlisting}[style=jsonstyle]
{
    "name": "get_stock_kline_history",
    "description": "Get historical K-line data for a stock, including daily/weekly/monthly K-lines, returns OHLC and volume data",
    "inputSchema": {
        "properties": {
            "stock_code": {"type": "string"},
            "period": {
                "type": "string", 
                "enum": ["daily", "weekly", "monthly"]
            },
            "start_date": {"type": "string"},
            "end_date": {"type": "string"}
        },
        "required": ["stock_code", "period"]
    },
    "outputSchema": {
        "properties": {
            "kline_data": {
                "type": "array", 
                "description": "K-line data array"
            }
        }
    }
}
\end{lstlisting}
\end{toolbox}

\begin{toolbox}[title={\textbf{API B: Calculate Technical Indicators}}]
\begin{lstlisting}[style=jsonstyle]
{
    "name": "calculate_technical_indicators",
    "description": "Calculate technical analysis indicators based on K-line data, such as MACD, KDJ, Bollinger Bands, etc. Note: Must first call get_stock_kline_history to obtain K-line data",
    "inputSchema": {
        "properties": {
            "kline_data": {
                "type": "array", 
                "description": "K-line data array"
            },
            "indicators": {
                "type": "array", 
                "description": "List of indicators to calculate"
            }
        },
        "required": ["kline_data", "indicators"]
    }
}
\end{lstlisting}
\end{toolbox}

\begin{toolbox}[title={\textbf{Extracted Relationship}}]
\begin{lstlisting}[style=jsonstyle]
{
    "head": "get_stock_kline_history",
    "relation": "direct_tool_dependency",
    "tail": "calculate_technical_indicators"
}
\end{lstlisting}
\end{toolbox}

\noindent\textbf{Explanation:} The description of \texttt{calculate\_technical\_indicators} explicitly states ``Must first call get\_stock\_kline\_history to obtain K-line data.'' Technical indicator calculation completely depends on raw K-line data; without K-line data, no technical analysis can be performed. This represents a mandatory dependency in the analysis workflow.

\paragraph{Indirect Tool Dependency}

An indirect tool dependency exists when a tool's execution can benefit from another tool, but is not mandatory. Calling the prerequisite tool provides more comprehensive information or aids decision-making, offering users a more complete analytical perspective without being required for the query to function.

\begin{toolbox}[title={\textbf{API A: Industry Money Flow Query}}]
\begin{lstlisting}[style=jsonstyle]
{
    "name": "get_industry_money_flow",
    "description": "Query capital inflow/outflow for industry sectors, including institutional and retail capital flows",
    "inputSchema": {
        "properties": {
            "date": {
                "type": "string", 
                "description": "Query date"
            },
            "top_n": {
                "type": "integer", 
                "description": "Return top N industries"
            }
        },
        "required": []
    },
    "outputSchema": {
        "properties": {
            "industry_list": {"type": "array"},
            "net_inflow_ranking": {"type": "array"}
        }
    }
}
\end{lstlisting}
\end{toolbox}

\begin{toolbox}[title={\textbf{API B: Industry Constituents Query}}]
\begin{lstlisting}[style=jsonstyle]
{
    "name": "get_industry_constituents",
    "description": "Query all constituent stocks and their weights for an industry sector",
    "inputSchema": {
        "properties": {
            "industry_code": {
                "type": "string", 
                "description": "Industry code"
            },
            "industry_name": {
                "type": "string", 
                "description": "Industry name, e.g., 'New Energy', 'Semiconductor'"
            }
        },
        "required": []
    }
}
\end{lstlisting}
\end{toolbox}

\begin{toolbox}[title={\textbf{Extracted Relationship}}]
\begin{lstlisting}[style=jsonstyle]
{
    "head": "get_industry_money_flow",
    "relation": "indirect_tool_dependency",
    "tail": "get_industry_constituents"
}
\end{lstlisting}
\end{toolbox}

\noindent\textbf{Explanation:} The API \texttt{get\_industry\allowbreak\_constituents} can independently complete queries; users can directly specify the industry name they want to explore (e.g., ``New Energy''). However, first calling \texttt{get\_industry\_money\_flow} to see which industries have the highest capital inflow can help users discover noteworthy trending industries before further querying constituent stocks. This represents an auxiliary relationship in the analysis chain.

\subsubsection{Relationship Priority Rules}

We adopt a four-level priority system, ranked from highest to lowest as follows:
\begin{enumerate}
    \item \texttt{direct\_parameter\_dependency} (Direct Parameter Dependency)
    \item \texttt{direct\_tool\_dependency} (Direct Tool Dependency)
    \item \texttt{indirect\_parameter\_dependency} (Indirect Parameter Dependency)
    \item \texttt{indirect\_tool\_dependency} (Indirect Tool Dependency)
\end{enumerate}

Parameter-level dependencies are more specific than tool-level dependencies, and direct dependencies are stronger than indirect dependencies. During annotation, when a \texttt{direct\_parameter\_dependency} has been identified between the same pair of APIs, we do not additionally add \texttt{direct\_tool\_dependency}, as the former already implicitly contains the semantics of the latter, thereby avoiding redundant annotations.

\subsubsection{Vector Encoding Strategy}

Taking \texttt{search\_company\_by\_name} as an example, our vector encoding strategy focuses on two core fields: tool name and tool description. Specifically, we extract the tool name ``search\_company\_by\_name'' and the description text ``Fuzzy search by company name keywords, returns matching company codes (company\_code),'' concatenate them, and feed the result into a pre-trained language model for encoding to obtain the semantic vector representation of this API:
\begin{equation}
    \mathbf{h}_i = \text{Encoder}(\text{Concat}(n_i, d_i))
\end{equation}
where $n_i$ denotes the tool name and $d_i$ denotes the tool description. This encoding approach captures the functional semantics of tools, making APIs with similar functions closer in vector space, providing a foundation for subsequent dependency relationship identification.

\subsection{Dataset Statistics and Distribution Analysis }
\label{app:dataset_distribution}
To ensure a rigorous evaluation of the model's tool-use capabilities, we constructed a dataset comprising \textbf{843 high-quality samples}. This dataset is designed to cover a wide spectrum of conversational complexities and tool execution patterns, preventing the model from overfitting to simple, single-step queries. We analyze the dataset characteristics from two perspectives: overall dimensional distribution and specific scenario combinations.

\paragraph{Overall Distribution Analysis}

Table~\ref{tab:dataset_stats} presents the dataset statistics across four key dimensions: Conversation Structure, Response Category, Tool Execution Pattern, and Candidate Tool Environment.

\begin{itemize}
    \item \textbf{Conversation Structure (Context Retention):} The dataset maintains a near-perfect equilibrium between \textit{Single-turn} (51.25\%) and \textit{Multi-turn} (48.75\%) interactions. This balance is crucial for assessing both the model's immediate instruction-following ability and its capacity to maintain context and handle state dependencies over prolonged dialogues.
    
    \item \textbf{Response Category (Intent Discrimination):} While the majority of samples require \textit{Tool Calls} (62.63\%), we deliberately included a significant proportion of non-tool responses (approx. 37\%). This includes \textit{Normal Replies}, \textit{Clarifications}, and \textit{No Tool Available} scenarios. This distribution tests the agent's "rejection capability"—the ability to discern when \textit{not} to invoke a tool, which is as critical as correct invocation.
    
    \item \textbf{Tool Execution Pattern (Planning Capability):} To challenge the agent's planning logic, the dataset goes beyond simple \textit{Single Calls} (26.33\%). A substantial portion involves complex patterns such as \textit{Parallel Calls} (24.32\%) and \textit{Serial Calls} (11.98\%), requiring the model to decompose tasks and manage dependencies between tool outputs.
    
    \item \textbf{Candidate Tool Environment (Retrieval Robustness):} Over 36\% of the samples are situated in a \textit{Multi-tool Context}, where the model must select the correct tool from multiple distractors, evaluating its retrieval accuracy and robustness against noise.
\end{itemize}

\paragraph{Scenario Combination and Balance}

We further categorize the dataset into 16 distinct scenarios derived from the combination of the four dimensions mentioned above. The detailed distribution is shown in Table~\ref{tab:dataset_stats}.

A distinguishing feature of our benchmark is the \textbf{uniform distribution} across these combinations. As illustrated in the table, the sample count for each scenario is strictly controlled between 50 and 63. 

This uniformity ensures that the evaluation metrics are not skewed by the prevalence of simple queries. Even the most complex scenarios—such as \textit{Multi-turn dialogues requiring Serial Calls in a Multi-tool context}—have sufficient representation (50 samples, 5.93\%). This design allows for a reliable measurement of the model's robustness across varying levels of difficulty, ensuring that high scores reflect genuine capability rather than overfitting to easy patterns.

\begin{table*}[t!]
\centering
\renewcommand{\arraystretch}{1.1}
\setlength{\tabcolsep}{6pt}
\small
\begin{tabular}{llllrr}
\toprule
\textbf{Round Type} & \textbf{Reply Type} & \textbf{Execution Pattern} & \textbf{Tool Context} & \textbf{Count} & \textbf{Pct. (\%)} \\
\midrule
Single-turn & Tool Call & Single & Single-tool & 63 & 7.47 \\
Single-turn & Normal Reply & Null & Null & 56 & 6.64 \\
Single-turn & Tool Call & Single & Multi-tool & 55 & 6.52 \\
Multi-turn & Tool Call & Single & Single-tool & 54 & 6.41 \\
Single-turn & No Tool Reply & Null & Null & 53 & 6.29 \\
\midrule
Multi-turn & Normal Reply & Null & Null & 52 & 6.17 \\
Multi-turn & Follow-up & Null & Null & 52 & 6.17 \\
Single-turn & Follow-up & Null & Null & 52 & 6.17 \\
Multi-turn & Tool Call & Parallel & Multi-tool & 52 & 6.17 \\
\midrule
Single-turn & Tool Call & Parallel & Multi-tool & 51 & 6.05 \\
Multi-turn & Tool Call & Parallel & Single-tool & 51 & 6.05 \\
Single-turn & Tool Call & Serial & Multi-tool & 51 & 6.05 \\
Single-turn & Tool Call & Parallel & Single-tool & 51 & 6.05 \\
\midrule
Multi-turn & No Tool Reply & Null & Null & 50 & 5.93 \\
Multi-turn & Tool Call & Single & Multi-tool & 50 & 5.93 \\
Multi-turn & Tool Call & Serial & Multi-tool & 50 & 5.93 \\
\bottomrule
\textbf{Total} & & & & \textbf{843} & \textbf{100.00} \\
\bottomrule
\end{tabular}
\caption{Detailed distribution of the 16 combined scenarios. The dataset maintains a balanced distribution across these combinations (50--63 samples each) to ensure comprehensive evaluation coverage without bias toward specific query types.}
\label{tab:dataset_stats}
\end{table*}

\subsection{API System Architecture}
\label{app:api_system}

To construct a comprehensive and logically structured Financial API System, we employed a hierarchical, human-in-the-loop taxonomy construction pipeline. Rather than relying on static categorization, we utilized LLMs to iteratively cluster APIs in a coarse-to-fine manner. The process first derived broad Level-1 domains (e.g., Account Management), which were rigorously verified by human experts. Subsequently, we generated granular Level-2 functional tags to capture specific capabilities, ensuring a precise mapping of the financial lifecycle. As detailed in Table~\ref{tab:api_system}, this architecture spans from core infrastructure like \textit{Payment \& Settlement} to advanced analytics such as \textit{Risk Management} and \textit{ESG Finance}, allowing developers to seamlessly integrate modular financial capabilities.

\onecolumn
\renewcommand{\arraystretch}{1.3} %

\begin{longtable}{p{0.95\textwidth}}
\toprule
\multicolumn{1}{c}{\textbf{Category \& Functions}} \\
\midrule
\endfirsthead

\toprule
\multicolumn{1}{c}{\textbf{Category \& Functions}} \\
\midrule
\endhead

\endfoot

\bottomrule
\caption{Financial API System Overview} \label{tab:api_system} \\

\endlastfoot

\textbf{Payment \& Settlement} \\*
Payment transaction processing, Payment status query and monitoring, Settlement, clearing and reconciliation, Cross-border payments, Payment channel management \\
\midrule

\textbf{Insurance Services} \\*
Underwriting and quoting, Policy lifecycle management, Claims services and processing, Reinsurance and risk transfer, Insurance product catalog and configuration management, Insurance operations analysis and monitoring \\
\midrule

\textbf{Account \& User Management} \\*
User registration and account opening, KYC and compliance management, User profile management, Identity authentication and authorization, Permissions and role management, Account lifecycle management, Account information query, Account security management \\
\midrule

\textbf{Custody \& Trust Services} \\*
Asset custody and depository, Trade clearing and settlement, Valuation accounting and reporting, Trust service management, Asset transfer \\
\midrule

\textbf{Customer Service \& Communication} \\*
Customer support and contact channels, Ticket management, Notification and reminder center, FAQ and knowledge base management, User feedback and analysis, Online chat and session management, Resource and document support, Personalized service and user behavior analysis, Conversation parsing and intent recognition, Complaints and legal support \\
\midrule

\textbf{Reporting Services} \\*
Report generation, Report type management, Report metadata management, Report query, Report push, Report export and conversion, Report validation and parsing \\
\midrule

\textbf{Developer Support} \\*
SDKs and development tools, Error diagnosis and debugging support, Sandbox and integration testing environment, API runtime monitoring and performance analysis, API documentation and reference, Security and compliance integration support, Integration verification and platform compatibility \\
\midrule

\textbf{Trading \& Execution} \\*
Algorithmic trading, Trading simulation and backtesting, Trading cost management, Trading strategy and optimization, Arbitrage and hedging strategies, Trading rules and mechanism configuration, Trading session management, Trade monitoring and alerts, Liquidity management, Margin and financing services, Order flow analysis, Security screening and matching, Risk management tools, Trade execution services, Order management system, Derivatives trading support, Counterparty services \\
\midrule

\textbf{Education \& Training} \\*
Personalized learning recommendations and paths, Learning assessment and testing, Financial knowledge education, Investment education and strategies, Case studies and practical analysis, API and operation tutorials, Simulated trading systems, Policy, regulation and fee education, Market dynamics and trading rules education, Education resource management, Financial terminology and historical literature, Certification and examinations, Interactive Q\&A platforms, Visualization tools and support, Risk management education, Financial planning and savings education, Financial operation training and tutorials, Professional skill training \\
\midrule

\textbf{Market Data Services} \\*
Real-time market data, Historical market data, Tick and transaction-level data, Candlestick data, Market depth and order book, Technical indicator data, Volume and position data, Asset metadata query, Stock market data, Indices and constituents, Sector and industry data, ETF and fund data, Bond and fixed income data, Derivative market data, Commodities data, Forex and exchange rate data, Cryptocurrency data, REITs market data, Market statistics and rankings, Fund flow analysis, Market sentiment indicators, Volatility data, Margin financing data, Market status and rules, Trading session and calendar, Market events and special quotes, Data quality and validation services \\
\midrule

\textbf{Data Management \& Governance} \\*
Data cleaning and standardization, Data parsing and extraction, Data ingestion and integration, Data consolidation and aggregation, Data classification and labeling, Data query and retrieval, Temporal data management, Geographic information management, Data metadata management, Data quality management, Data lineage and traceability, Data security and compliance, Data validation and auditing, Data monitoring and anomaly detection, Data governance rules and configuration, Data lifecycle management, Data source management, Data subscription and update management, Data governance assessment, Entity recognition and extraction, Data analysis and mining \\
\midrule

\textbf{Asset Lifecycle Management} \\*
Asset verification and reporting, Asset monitoring and alerts, Asset query and screening, Asset terms and rules query, Asset issuance and registration, Corporate actions processing, Asset analysis and risk assessment, Position query and management, Asset information management, Asset change notifications, Delisting management, Asset custody and depository, Asset movement and settlement, Asset valuation and performance \\
\midrule

\textbf{Financial Tools \& Calculation} \\*
Investment calculators, Financial model computation, Investment return calculation, Basic financial calculations, Trading cost calculation, Business day calculations, Numeric standardization and conversion, Deposit interest calculation, Portfolio and risk management, Growth rate calculation, Financial health and risk assessment, Fund return calculation, Derivatives pricing calculator, Risk premium calculation, Compound interest calculation, Debt calculation, Technical indicator computation, Stock return calculation, Retirement planning, Compensation calculation, Price fluctuation calculation, Internal rate of return calculation, Derivatives pricing and calculation, Hedging calculation, Asset valuation calculation, Portfolio analysis, Stock trend calculation, Education fund planning, Expense and cost analysis, Option pricing and hedging, Bond calculations, Trading fee calculators, Margin and leverage calculation, Date and time tools, Stock pricing calculation, Fund calculation, Deposit yield calculation, Return rate calculation, Financial health assessment, Date conversion tools, Investment return evaluation, Expense and cost calculation, Bond duration calculation, Loan calculation, Mortgage calculation, Price calculation, Retirement and education planning, Probability and statistics calculations, Leverage calculation, Financial forecasting calculators, Stock-related calculations, Financial ratio calculations, Interest rate calculation, Household financial health assessment, Financial indicators calculation, Investment return and risk analysis, Dividend calculation, Date calculator, Subsidy calculation, Project financial evaluation, Cash flow and investment analysis, Investment yield calculation, Implicit volatility calculation, Volatility calculation, Derivatives pricing and risk, Inflation-adjusted calculations, Cash flow and fund calculation, Unit conversion calculation, Risk and return assessment, Economic indicators calculation, Margin calculation, Savings calculation, Financial planning calculation, Monte Carlo simulation, Cost of capital calculation, Derivatives risk management, Income calculation, Fund calculators, Financial forecasting, Derivatives pricing and hedging, Interest calculation, Option pricing, Cash flow analysis, Stock and equity calculations, Economic indicator calculators, Stock pricing and calculation, Bond return calculation, Numeric computation tools, Deposit rate calculators, Interest and yield calculation, Bond pricing calculators, Bond yield calculation, Net asset value calculation, Yield curve calculation, Financial statistics calculation, Financial indicators and model calculation, Education fund computation, Financial leverage calculation, Probability distribution computation, Bond pricing, Net present value calculation, Investment return rate calculation, Tax and fee calculation, Deposit and interest calculation, Cash flow modeling, Cost optimization calculation, Tax calculation, Statistical indicator calculation, Exchange rate calculation, Derivatives pricing, Retirement planning calculation, Emergency expense calculation, Forex and currency calculation, Cash flow forecasting, Financial model simulation, Cost and expense calculation, Time and unit conversion, Technical indicator analysis \\
\midrule

\textbf{Investment \& Portfolio Management} \\*
Specific asset class analysis, Risk analysis and management, Investment strategy generation and analysis, Portfolio rebalancing and monitoring, Trading execution support, Performance analysis, Product analysis and screening, Cost optimization, Investor profiling and preference analysis, Quantitative analysis, Investment simulation and backtesting, Portfolio construction and optimization, Market and industry analysis, Asset allocation and planning \\
\midrule

\textbf{Financial \& Tax Services} \\*
Tax monitoring and alerts, Policy query and matching, Tax regulation query and compliance, Government subsidies and aid query, Tax calculation, Tax risk management, Tax filing support, Financial information management, Tax analysis and reporting, Invoice management, Tax policy query, Tax information query, Income and expense analysis, Tax consultation and advice, Cross-border tax support, Investment tax support, Accounting support, Tax simulation, Tax optimization and planning, Social insurance and housing fund query, Asset tax management, Accounting and financial verification, Tax practice support, Tax incentive management \\
\midrule

\textbf{Research \& Analysis} \\*
Valuation and pricing analysis, Company fundamental analysis, Technology and innovation analysis, Macro market and policy analysis, Industry research, Supply chain and industry chain analysis, Fixed income securities analysis, Financial modeling and valuation, Research report and interpretation, Credit rating and risk analysis, Event and case study, Policy and regulatory analysis, Market sentiment and behavior analysis, Data query and retrieval, Investment strategy and quantitative analysis \\
\midrule

\textbf{ESG \& Sustainable Finance} \\*
Technology and economic analysis, Social and governance assessment, Corporate governance information, Green projects and certification, Clean technology certification, Policy and compliance, Governance analysis, Project evaluation and feasibility analysis, Energy efficiency and optimization, Green bonds, Green finance analysis, ESG data and assessment, Industry benchmarks and comparison, ESG trends and analysis, Carbon and emissions reduction, Green financing and transition, ESG risk monitoring, Supply chain sustainability, Sustainable development reporting, Sustainable investment, Governance risk assessment, Green subsidies and incentives, Infrastructure and energy projects, ESG rating and scoring, Environmental and social risk monitoring, Sustainable policies and frameworks, ESG negative event monitoring, Environmental benefits and verification \\
\midrule

\textbf{Lending \& Financing} \\*
Loan approval status query, Financing conditions and channels query, Loan data statistics, Loan quality and risk analysis, Loan contract template management, Financing inquiry, Risk pricing, Financing progress monitoring, Loan due diligence, Credit limit management, Financing cost calculation, Financing feasibility assessment, Repayment plan, Loan contract management, Loan product query, Interest rate query and analysis, Loan query, Repayment risk assessment, Financing plan, Credit evaluation, Loan structure analysis, Collateral management, Loan portfolio analysis, Loan report generation, Loan application and approval, Debt management, Loan product recommendation, Policy query, Loan risk model, Loan customer profiling, Loan automated approval, Loan terms query \\
\midrule

\textbf{Risk Management} \\*
Risk warning and event monitoring, Risk model construction, Risk factor analysis, Industry risk monitoring, Risk information analysis, Risk alerts and monitoring, Asset risk assessment, Risk quantification and analysis, Risk identification and alerts, Portfolio risk management, Case study and retrieval, Risk management strategies, Quantitative model risk analysis, Risk monitoring configuration, Liquidity risk analysis, Risk simulation, Comprehensive risk analysis, Operational risk analysis, Risk hedging strategies, Risk models and backtesting, Stress testing and scenario simulation, Risk models and reporting, Risk parameters and threshold management, Risk simulation and reporting, Risk control measures, Policy and macro risk assessment, Risk indicator calculation, Dynamic risk management, Risk response and strategies, Risk strategy and hedging, Risk event analysis, Policy risk warning, Concentration risk monitoring, Risk weighting calculation, Historical risk data analysis, Risk type identification and monitoring, Risk hedge assessment, Financial risk monitoring, Credit risk assessment, Market risk monitoring and alert, Policy and external risk monitoring, Risk parameter analysis, Risk exposure and assessment, Scenario risk assessment, Market risk analysis, Risk analysis, Risk warning and analysis, Risk appetite and tolerance, Risk model validation, Real-time monitoring alerts, Risk strategy analysis, Risk factor monitoring and alerts, Risk indicator query, Industry and regional risk monitoring, User risk assessment, Risk monitoring dashboard, Risk factor identification and quantification, Risk mitigation strategies, Risk monitoring and alerts, Cash flow risk management, Industry and macroeconomic risk monitoring, Operational risk assessment, Risk monitoring and alarms, Risk hedging and strategies, Risk signal identification, Risk event identification and assessment, Risk identification, Risk factor identification and monitoring, Comprehensive risk monitoring, Credit risk monitoring, Systemic risk assessment, Supply chain risk assessment, Collateral and guarantee assessment, Risk reporting and visualization, Risk control and mitigation, Risk simulation and scenario analysis, Liability and liquidity risk monitoring, Risk strategy management, Risk monitoring, Collateral valuation and monitoring, Risk appetite and assessment, Risk evaluation and analysis, Risk mitigation and hedging, Default probability analysis, Debt risk analysis, Enterprise risk assessment, Sensitivity analysis \\
\midrule

\textbf{Compliance \& Regulation} \\*
Data privacy and compliance, Rules and policy management, Rule query and analysis, Legal and regulatory search, Regulatory reporting and submission, KYC (Know Your Customer), Corporate governance, Enterprise compliance assessment, Compliance information query, Regulatory penalties and litigation query, Regulatory approval, Compliance guidelines and standards, Enterprise compliance monitoring, Compliance monitoring and alert, Shareholder and related-party identification, Qualification verification, Compliance status verification, Regulation search and compliance check, Compliance record management, Audit and historical record, Trading rules and verification, Compliance risk assessment, Policy and rule interpretation, Document validation and generation, Regulation and policy monitoring, AML (Anti-Money Laundering), Regulation search and compliance evaluation, Regulatory inquiry and penalty scan, Regulation compliance check, Compliance review and verification, Compliance inspection process, Compliance inspection and assessment, Regulatory tracking, Information disclosure and supervision, Major event identification, Risk disclosure and governance information query, Compliance violation query, Legal and judicial support, Audit and financial report verification, Regulatory compliance information query, Judicial and legal compliance, Policy compliance assessment, Trading monitoring, Internal control and audit, Investor suitability, Compliance monitoring, Policy and regulation search and analysis, Regulatory policy monitoring, Regulatory information query, Regulatory sandbox \\

\end{longtable}

\clearpage
\twocolumn

\subsection{Synthesizing Dialogue Record}
\label{app:synthesis_examples}
\begin{quote}
\textit{"Autowell's dominance in the photovoltaic equipment sector is solid, but are there any potential synergies between its lithium-ion energy storage business and autonomous driving technology?"}
\end{quote}

\begin{userbox}[title={\faIdCard\ \textbf{High Net Worth Individual (HNWI) Persona}}]

    \begin{minipage}[t]{0.48\textwidth}
        \textcolor{userframe}{\large\textbf{Basic Profile}}
        \par\noindent\rule{\textwidth}{0.4pt} \vspace{0.5em}
        
        \profileitem{\faBirthdayCake}{Age}{45 Years Old}
        \profileitem{\faMars}{Gender}{Male}
        \profileitem{\faRing}{Marital Status}{Married}
        \profileitem{\faGraduationCap}{Education}{M.S. in Finance}
        \profileitem{\faChild}{Children}{2}
        \profileitem{\faBriefcase}{Occupation}{Senior Investment Director}
        \profileitem{\faLayerGroup}{Social Class}{Upper-Middle Class}
        \profileitem{\faBrain}{Personality}{Cautious yet Curious}
    \end{minipage}%
    \hfill
    \begin{minipage}[t]{0.48\textwidth}
        \textcolor{userframe}{\large\textbf{Financial Profile}}
        \par\noindent\rule{\textwidth}{0.4pt} \vspace{0.5em}
        
        \profileitem{\faWallet}{Economic Status}{Affluent}
        \profileitem{\faChartLine}{Financial Literacy}{Expert Level}
        \profileitem{\faLayerGroup}{Risk Tolerance}{High}
        \profileitem{\faChess}{Planning Awareness}{Strong \& Strategic}
        \profileitem{\faHistory}{Experience}{Extensive / Institutional}
        
        \vspace{0.5em}
        \textbf{\faLightbulb\ Investment Preferences:}
        \begin{itemize}
            \setlength\itemsep{-0.2em}
            \small
            \item Early-stage Tech \& AI
            \item Blockchain Technology
            \item Autonomous Driving
        \end{itemize}
    \end{minipage}

\end{userbox}

\begin{userbox}
How large is Autowell's lithium-ion energy storage business now? Are the main clients companies like Canadian Solar or Trina Solar? Is it considered a core business within the company?
\end{userbox}

\begin{modelbox}
\textit{Thinking: The user is asking about market scale, specific clients, and strategic importance. I need to query financial metrics and client data.}

\begin{toolbox}
\begin{lstlisting}[style=jsonstyle]
[get_new_energy_business(company_name='Autowell', metrics=['New Energy Order Volume', 'Energy Storage Revenue Share']), get_client_order_trends(company_name='Autowell')]
\end{lstlisting}
\tcblower %
\begin{lstlisting}[style=jsonstyle]
[
    "{"New Energy Order Volume": 1.85 billion, "Energy Storage Revenue Share": 37.2%
    "{"major_clients": [{"client_name": "CATL", "order_size_million": 280, "trend": "increasing", "order_share_percent": 42}, {"client_name": "BYD", "order_size_million": 195, "trend": "stable", "order_share_percent": 28}], "overall_demand_outlook": "strong growth driven by expansion in global energy storage projects"}"
]
\end{lstlisting}
\end{toolbox}

Autowell's lithium-ion energy storage business has developed to a significant scale. Its new energy-related order volume has reached \textbf{1.85 billion RMB}, and energy storage revenue now accounts for \textbf{37.2\%} of the company's total revenue, making it one of the core business segments. 

Regarding clients, the main partners include \textbf{CATL} and \textbf{BYD}. Orders from CATL account for 42\% and show a continuous growth trend. Overall, the company is deeply bound to top domestic battery enterprises and is benefiting from strong global demand for energy storage, placing this segment in a rapid development phase.
\end{modelbox}

\begin{userbox}
What is the specific connection between Autowell's PACK production line and battery systems in the autonomous driving sector? Where can it be applied in the future?
\end{userbox}

\begin{modelbox}
\begin{toolbox}
\begin{lstlisting}[style=jsonstyle]
[query_NEV_related_business(company_identifier='688516.SH', keywords=['PACK Production Line', 'Battery System', 'Autonomous Driving'])]
\end{lstlisting}
\tcblower
\begin{lstlisting}[style=jsonstyle]
{
  "company_identifier": "688516.SH",
  "related_business_areas": [
    {
      "business_area": "PACK Production Line",
      "details": "Subsidiary has built a fully automatic PACK line with 5GWh annual capacity; clients include mainstream EV manufacturers."
    },
    {
      "business_area": "Battery System",
      "details": "Self-developed High-Voltage Battery System passed automotive-grade safety certification; integrates liquid cooling & BMS."
    }
  ],
  "no_match_keywords": ["Autonomous Driving"],
  "notes": "No R&D projects, products, or patents directly related to 'Autonomous Driving' were found."
}
\end{lstlisting}
\end{toolbox}

There is currently \textbf{no direct technical connection} between Autowell's PACK production line and battery systems specifically for the autonomous driving sector. 

The company's existing PACK lines are primarily used for producing power battery systems for New Energy Vehicles (NEVs), with an annual capacity of 5GWh. Its self-developed high-voltage battery systems integrate liquid cooling and intelligent BMS modules, falling under the category of automotive-grade power systems. However, based on current information, Autowell has not engaged in R\&D for battery or energy management technologies specifically for autonomous driving systems. In the future, as vehicle electrification and intelligence merge, its expertise in BMS and thermal management might offer synergistic support, but the current focus remains on power battery manufacturing equipment.
\end{modelbox}

\begin{userbox}
So, regarding the PACK line in power batteries, what exactly are the technical advantages? Where are the points that are difficult for others to replicate?
\end{userbox}

\begin{modelbox}
\begin{toolbox}
\begin{lstlisting}[style=jsonstyle]
[scan_battery_patents(applicant_keyword='Autowell')]
\end{lstlisting}
\tcblower
\begin{lstlisting}[style=jsonstyle]
[
  "{
    "patent_analysis": [
        {
            "title": "Smart Battery Module & Management Method for Storage Systems", "core_innovation": "Integrated Thermal Management & Dynamic Balancing Algorithm"
        }, 
        {
            "title": "Digital Twin-based PACK Line Monitoring System", 
            "core_innovation": 
            "Virtual-Real Linkage Fault Prediction & Capacity Optimization Model"
        }], 
    "technology_barrier_score": 82, 
    "trend_analysis": "Patent applications grew by 35%
    }"
]
\end{lstlisting}
\end{toolbox}

Autowell's technical advantage in the PACK production line lies in its \textbf{"Software-Hardware Integrated" smart manufacturing system}. The core barriers to replication include:

1.  **Digital Twin Technology**: Their patented "Digital Twin-based PACK Line Monitoring System" builds a model for real-time simulation and fault prediction. This relies on long-term process data accumulation, making it hard to copy.
2.  **Deep Integration**: The PACK line deeply fuses intelligent BMS with thermal management (e.g., dynamic balancing algorithms), ensuring superior safety and consistency.
3.  **Continuous Innovation**: With a 35\% annual growth in patents, their focus on digitalization creates a constantly evolving moat that goes far beyond simple equipment automation.
\end{modelbox}

\begin{userbox}
What will drive the continued growth of Autowell's storage business? Any new market plans?
\end{userbox}

\begin{modelbox}
\begin{toolbox}
\begin{lstlisting}[style=jsonstyle]
[track_enterprise_capacity(companies=['Autowell'], metric_type='Capacity Planning')]
[get_new_energy_application(stock_code='688516.SH', industry_keywords=['Energy Storage', 'Overseas Market'])]
\end{lstlisting}
\tcblower
\begin{lstlisting}[style=jsonstyle]
[
  "{"capacity_planning": 1200, "rank_comparison": "Top Tier"}",
  "{"applications": [{"case_name": "European Large-scale Storage Project", "description": "100MWh system for a German energy company."}, {"case_name": "SE Asia PACK Line Delivery", "description": "AutoPack Pro 5.0 solution delivered to Thailand/Vietnam."}], "summar\": {"total_overseas_revenue_share": 34.6}}"
]
\end{lstlisting}
\end{toolbox}

The continued growth of Autowell's storage business relies on three core drivers:

1.  **Capacity Expansion**: Planning for \textbf{1,200 MWh} of capacity places them at the forefront of the industry, guaranteeing order fulfillment.
2.  **Technical Barriers**: Continuous iteration of Digital Twin monitoring and intelligent BMS technologies.
3.  **Global Layout**: The company has successfully entered European and Southeast Asian markets (e.g., 100MWh project in Germany, PACK lines in Thailand/Vietnam). Overseas revenue now accounts for \textbf{34.6\%} of the segment. Future growth will come from deepening these global partnerships and leveraging ties with leaders like CATL to export smart manufacturing solutions.
\end{modelbox}

\subsection{Process Supervision}
\label{app:global_control_examples}
This section illustrates the multi-stage supervision framework that ensures dialogue quality and agent behavior correctness during the synthesis process. The pipeline begins with an \textbf{Initial Dialogue Plan} that outlines the intended trajectory of the conversation. As the user issues queries, a \textbf{Confusion Test} evaluates whether the generated user message authentically matches the persona's knowledge level, language style, and behavioral traits—distinguishing genuine queries from implausible or test-like inputs. A \textbf{Quality Test} then verifies compliance with the dialogue plan, behavioral consistency, persona alignment, and formatting rules. Following the assistant's response, a \textbf{Process Analysis} module audits the reasoning logic, tool selection, parameter correctness, and response appropriateness. Finally, the \textbf{Dialogue Planning} mechanism checks whether the current plan remains valid or requires updates based on conversational dynamics. This layered supervision architecture guarantees that synthesized dialogues are not only realistic and persona-consistent but also logically sound and quality-controlled at every step.

\vspace{1em}
\noindent\textbf{Step 1: Initial Dialogue Planning.} The supervision process begins with the Global Agent establishing a structured dialogue plan. This plan defines the expected trajectory of the conversation, specifying the topic and intent for each round. The plan serves as a reference baseline for subsequent quality checks and ensures the synthesized dialogue follows a coherent, goal-oriented structure.

\begin{agentbox}[title={\faClipboardList\ Global Agent: Dialogue Planning}]
    \textbf{Initial Plan:}
    \begin{itemize}
        \item \textbf{Round 1:} Inquire about the specific credit risk situation of Fund 019667.
        \item \textbf{Round 2:} Follow up asking for specific countermeasures or suggestions regarding the fund's current credit risk status.
    \end{itemize}
\end{agentbox}

\vspace{1em}
\noindent\textbf{Step 2: User Query Generation.} Based on the dialogue plan, the User Agent generates the first query. This query is designed to align with both the plan's intent and the user persona's characteristics.

\begin{userbox}
    How is the credit risk of Fund 019667 specifically?
\end{userbox}

\vspace{1em}
\vspace{1em}
\noindent\textbf{Step 3: Confusion Test (Round 1).} The Global Agent receives both queries and evaluates their authenticity against the user persona.

\begin{agentbox}[title={\faUserSecret\ Global Agent: Confusion Test}]
    \textbf{Analysis Process:}

    Based on the guidelines, I need to determine the likelihood of each query.

    \textbf{Persona Compatibility:}
    \begin{itemize}
        \item \textbf{Text A:} Directly mentions the specific fund code "019667" explicitly held in the persona, and inquires about its "credit risk". This is highly compatible with the user's "lack of professional financial knowledge, relying on...online forums" and the anxious mentality of "worrying about losses affecting expenses". A real user would directly mention assets they hold.
        \item \textbf{Text B:} Asks a general, methodological question: "How to measure the default risk of credit funds?" This question is relatively abstract and academic. For a user who "lacks professional financial knowledge", while asking such a generalized question is not entirely impossible, its fit with the persona is far lower than Text A.
    \end{itemize}

    \textbf{Language Style \& Authenticity:}
    \begin{itemize}
        \item \textbf{Text A:} The language is very colloquial and natural. "What's the specific situation?" is a typical informal verbal questioning style, consistent with the expression habits of a non-professional seeking specific, easy-to-understand answers.
        \item \textbf{Text B:} The language is formal and structured, more like a standard question from a textbook or examination. The use of terms like "measure" and "default risk" is relatively standardized. While humans may also ask questions this way, its style leans more toward a "standard question", with less naturalness compared to Text A.
    \end{itemize}

    \textbf{AI Generation Traces:}
    \begin{itemize}
        \item \textbf{Text A:} No obvious AI traces. Its specificity and colloquial characteristics are more like a real person asking questions.
        \item \textbf{Text B:} Has slight but noteworthy AI or standardized test traces. Its questioning style is overly neat and generic, similar to a question designed to trigger a standard answer about "credit risk measurement indicators".
    \end{itemize}

    \textbf{Comprehensive Judgment:}
    \begin{itemize}
        \item Text A is highly consistent with the user persona in both content (mentioning specific holdings) and language style, making it a very likely and natural genuine user query.
        \item Although Text B is thematically relevant, its generality and relatively formal language style make its likelihood as a query from this specific user significantly lower than Text A. It appears more like a question designed to test model knowledge, rather than a question that an anxious individual investor holding a specific fund would ask.
        \item According to the guidelines "Likelihood > Typicality" and "Identify AI Characteristics", I believe \textbf{Text A and only Text A} is the genuine user-style query.
    \end{itemize}

    \textbf{Final Conclusion:} \framebox{A}
\end{agentbox}

\vspace{1em}
\noindent\textbf{Step 4: Quality Test (Round 1).} The Global Agent verifies whether the selected query complies with dialogue planning and persona requirements.

\begin{agentbox}[title={\faCheckCircle\ Global Agent: Quality Test}]
    \textbf{Analysis Process:}

    1. \textbf{Plan Consistency Analysis (Does the topic match?)}
    \begin{itemize}
        \item The dialogue history \texttt{<Dialogue\_History>} contains 0 \texttt{<user>} tags, so K=0. The current execution should correspond to Plan Item 1: "Inquire about the credit risk status or level of Fund 019667."
        \item The user query directly asks about the credit risk of Fund 019667, perfectly matching the planned first inquiry topic.
        \item Conclusion: \textcolor{teal}{[Plan Consistency: Compliant]}
    \end{itemize}

    2. \textbf{Behavior Consistency Analysis (Does the action match?)}
    \begin{itemize}
        \item The core verb in Plan Item 1 is "inquire". This means the user action should be a question-asking behavior.
        \item The user query is phrased as a direct question ("What's the specific situation?"), fully matching the expected questioning behavior.
        \item Conclusion: \textcolor{teal}{[Behavior Consistency: Compliant]}
    \end{itemize}

    3. \textbf{Persona \& Style Matching Analysis (Is the speaking manner appropriate?)}
    \begin{itemize}
        \item User persona: "cautious and pragmatic", "anxious", "Financial Knowledge: lacks professional financial knowledge", "relies on online forums and friends for advice".
        \item The user's wording is colloquial and direct ("What's the specific situation?"), consistent with everyday language habits of non-professionals seeking straightforward answers.
        \item Directly asking about a specific fund they hold reflects genuine concern about their investment, matching the anxious and cautious persona characteristics.
        \item Conclusion: \textcolor{teal}{[Persona \& Style Matching: Compliant]}
    \end{itemize}

    4. \textbf{Rule Compliance Analysis (Are there any rule violations?)}
    \begin{itemize}
        \item K=0, so greeting rules for subsequent turns do not apply to this first query.
        \item The sentence is concise and colloquial, not lengthy or overly formal.
        \item The query focuses on one specific topic: the credit risk of Fund 019667, without mixing unrelated issues.
        \item Conclusion: \textcolor{teal}{[Rule Compliance: Compliant]}
    \end{itemize}

    5. \textbf{Final Conclusion:}
    \begin{itemize}
        \item All four checks passed.
    \end{itemize}
    \framebox{True}
\end{agentbox}

\vspace{1em}
\noindent\textbf{Step 5: Assistant Response Generation.} The Assistant Agent processes the user query and generates a response. In this case, the assistant retrieves relevant data through tool calls and provides a comprehensive answer about the fund's credit risk profile.

\begin{modelbox}
    Regarding the credit risk of Fund 019667, the latest data shows a well-structured credit rating distribution across its bond holdings: AAA-rated bonds account for 25.5\%, AA+ for 30.0\%, AA for 20.0\%, AA- for 15.0\%, and A-rated for 9.5\%. The portfolio is predominantly composed of high-credit-quality bonds, indicating a low level of credit risk.

    Moreover, the fund has maintained a clean credit history with no defaults since inception. Its investment strategy emphasizes proactive credit risk management, focusing on high-rated bonds issued by state-owned enterprises and high-quality local government financing platforms. The fund strictly limits exposure to any single issuer to no more than 5\% of the portfolio. It also conducts regular stress tests and monitors credit spreads closely. For bonds showing early signs of risk, the fund implements timely warnings and reduces positions in advance.

    Overall, Fund 019667 exhibits low credit risk and operates under a robust risk management framework.
\end{modelbox}

\vspace{1em}
\noindent\textbf{Step 6: Process Analysis.} The Process Analysis module audits the assistant's response generation process. It verifies the logic flow, tool selection appropriateness, parameter correctness, and overall response quality. This ensures the assistant's behavior is both technically sound and contextually appropriate.

\begin{agentbox}[title={\faCogs\ Global Agent: Process Analysis}]
    \textbf{Assessment Analysis:}
    
    \textbf{Step 1: Logic Flow Check:}
    Last message is User. \texttt{ASSISTANT\_ANSWER\_TO\_REVIEW} Action is \texttt{choice 1} (Tool Call). This is a logical starting point.
    
    \textbf{Step 2: General Check:}
    \texttt{Thought} details the analysis of available tools and selects \texttt{query\_fund\_credit\_risk}. Consistency is good.
    
    \textbf{Step 3: Deep Check (Action: choice 1):}
    \begin{itemize}
        \item \textbf{Necessity:} User asks for specific data on "019667". Tool call is necessary.
        \item \textbf{Tool Selection:} \texttt{query\_fund\_credit\_risk} is selected. Output includes "credit rating distribution" and "default records", matching the user's request perfectly.
        \item \textbf{Parameter Completeness:} \texttt{fund\_identifier='019667'} is provided. Correct type (string) and format.
    \end{itemize}
    
    \textbf{Final Conclusion:} Logic is correct, execution is optimal.
    
    \textbf{Result:} \framebox{True}
\end{agentbox}

\vspace{1em}
\noindent\textbf{Step 7: Dialogue Planning Check (No Update Needed).} After Round 1 completes successfully, the Global Agent evaluates whether the dialogue plan requires modification. Since the conversation is progressing as expected and the assistant provided a complete answer without requesting additional information, the original plan remains valid. No updates are necessary.

\begin{agentbox}[title={\faClipboardList\ Global Agent: Dialogue Planning}]
    \texttt{<thought>}\\
    1. \textbf{Progress Check:} Round 1 completed successfully. Assistant provided comprehensive credit risk information.\\
    2. \textbf{Plan Evaluation:} The original Round 2 plan ("Follow up on countermeasures") remains appropriate as the next step.\\
    3. \textbf{Update Decision:} No modification needed. Plan continues as designed.\\
    \texttt{</thought>}
    
    \texttt{<completion\_status>} 
    
    \framebox{False}
    
    \texttt{</completion\_status>}
    
    \textbf{Current Plan (Unchanged):}
    \begin{itemize}
        \item \textbf{Round 1:} Inquire about 019667 credit risk details. \textcolor{teal}{[Completed]}
        \item \textbf{Round 2:} Follow up on countermeasures regarding low credit risk.
    \end{itemize}
\end{agentbox}

\vspace{1em}
\noindent\textbf{Step 8: Second Round User Query.} Following the plan, the User Agent generates a follow-up query asking for operational advice based on the low credit risk assessment.

\begin{userbox}
    Since the credit risk is low, how should I operate now? Should I continue holding or add to the position? Give me some advice.
\end{userbox}

\vspace{1em}
\noindent\textbf{Step 9: Confusion Test \& Quality Test (Round 2).} The same supervision checks are applied to the second user query. Both tests perform rigorous analysis to ensure the query's authenticity and compliance with the dialogue plan.

\begin{agentbox}[title={\faUserSecret\ Global Agent: Confusion Test}]
    \textbf{Analysis Process:}

    1. \textbf{Persona Anchoring:} The user is a 45-year-old male IT project manager, urban middle-class, with heavy family responsibilities, cautious and pragmatic personality, and anxious psychology. Limited financial knowledge, low risk tolerance, 10 years of investment experience but mainly relying on non-professional channels. Current assets include Fund 019667.

    2. \textbf{Analysis and Evaluation:}
    
    \textbf{Text A:} "Since the credit risk is low, how should I operate now? Should I continue holding or add to the position? Give me some advice."
    \begin{itemize}
        \item \textbf{Persona Compatibility:} Highly compatible. The query is based on the previous conclusion that "credit risk is low" (relevant to the bond fund the user holds), directly asking for specific operational advice ("holding" or "adding position"). The language is colloquial ("how should I operate", "give me some advice"), reflecting a non-professional questioning style, consistent with the persona of "relying on online forums/friends for advice". The question focuses on a single decision about current holdings, matching the "cautious and pragmatic" personality and "anxious" psychological state, as well as the need for direct advice despite "weak operability".
        \item \textbf{Language Style \& Authenticity:} Very authentic. The sentence is typical colloquial consultation, using informal expressions and a direct help-seeking tone like "give me some advice", fully consistent with natural conversation habits of humans (especially non-financial professionals).
        \item \textbf{AI Generation Traces:} No obvious AI traces. This is a coherent, natural, context-based human question.
    \end{itemize}

    \textbf{Text B:} "Can you summarize my previous questions and tell me if I should redeem now? And what should I do next? My goal is to further increase returns on a fixed-income basis, and my operability is weak. I'm an investment novice, so I also require easy-to-execute operations with relatively low risk."
    \begin{itemize}
        \item \textbf{Persona Compatibility:} Clear conflict exists. The user persona explicitly states "Investment Experience: 10 years", but Text B self-identifies as "an investment novice", which directly contradicts the core persona information. Although "weak operability" matches the persona, the self-description of "investment novice" severely undermines compatibility.
        \item \textbf{Language Style \& Authenticity:} Unnatural style. The sentence is lengthy, appearing to list all requirements and constraints at once ("goal...and...so I also require...and..."). This structure resembles deliberately organized instructions for an AI rather than the gradual or focused expression typical of real conversations. Real users would more likely break into separate sentences or supplement information through multiple interactions.
        \item \textbf{AI Generation Traces:} Some traces present. This way of packaging complex requirements (summary, specific operational advice, return goals, operability, risk preference) into one long sentence has the characteristic of "clearly explaining task background to AI", unlike the natural expression of an anxious, financially limited 45-year-old project manager in real-time conversation.
    \end{itemize}

    3. \textbf{Comprehensive Judgment:}
    \begin{itemize}
        \item Text A demonstrates high probability in persona compatibility, language style, and authenticity, with no AI traces.
        \item Text B has a hard conflict on key persona information (investment experience), and its language style appears contrived and unnatural, with traces of AI instructions.
        \item Therefore, based on the criterion that "there is one and only one genuine user-style query", Text A is the authentic user query while Text B is not.
    \end{itemize}

    \textbf{Final Conclusion:} \framebox{A}
\end{agentbox}

\begin{agentbox}[title={\faCheckCircle\ Global Agent: Quality Test}]
    \textbf{Analysis Process:}

    1. \textbf{Plan Consistency Analysis (Does the topic match?)}
    \begin{itemize}
        \item The dialogue history \texttt{<Dialogue\_History>} contains 1 \texttt{<user>} tag, so K=1. The current execution should correspond to Plan Item 2: "Follow up asking for specific countermeasures or suggestions regarding the fund's low credit risk status."
        \item The core topic of the user query is "how should I operate now", specifically proposing options of "continue holding or add to the position", directly seeking operational advice based on the low credit risk situation.
        \item This is fully consistent with the intent of Plan Item 2.
        \item Conclusion: \textcolor{teal}{[Plan Consistency: Compliant]}
    \end{itemize}

    2. \textbf{Behavior Consistency Analysis (Does the action match?)}
    \begin{itemize}
        \item The core verb in Plan Item 2 is "follow up...specific countermeasures or suggestions". This means the user action should be "asking questions" or "requesting advice".
        \item The user query contains explicit question sentences ("how should I operate now?") and advice-requesting expressions ("give me some advice"), fully matching the preset action of "following up for suggestions".
        \item Conclusion: \textcolor{teal}{[Behavior Consistency: Compliant]}
    \end{itemize}

    3. \textbf{Persona \& Style Matching Analysis (Is the speaking manner appropriate?)}
    \begin{itemize}
        \item User persona: "cautious and pragmatic", "anxious", "Financial Knowledge: lacks professional financial knowledge".
        \item The user's wording is very colloquial ("how should I operate", "add to the position", "give me some advice"), consistent with everyday language habits of non-professionals.
        \item After learning that the risk is low, immediately seeking specific operational advice reflects a "pragmatic" personality. Meanwhile, directly requesting "give me some advice" also reflects dependence on investment decisions and a certain anxious mentality, matching the persona.
        \item Conclusion: \textcolor{teal}{[Persona \& Style Matching: Compliant]}
    \end{itemize}

    4. \textbf{Rule Compliance Analysis (Are there any rule violations?)}
    \begin{itemize}
        \item K=1 (>0), the query does not contain any greetings (e.g., "hello") or pleasantries (e.g., "please", "thank you").
        \item The sentence is short and colloquial, not lengthy or overly formal.
        \item The query content focuses on one core point: "operational advice based on current credit risk status", without attempting to merge other unrelated topics (such as inquiring about other funds or macro markets).
        \item Conclusion: \textcolor{teal}{[Rule Compliance: Compliant]}
    \end{itemize}

    5. \textbf{Final Conclusion:}
    \begin{itemize}
        \item All four checks passed.
    \end{itemize}
    \framebox{True}
\end{agentbox}

\vspace{1em}
\noindent\textbf{Step 10: Assistant Response with Clarification Request.} The Assistant Agent determines that providing personalized investment advice requires additional information about the user's risk tolerance, investment goals, and position adjustment intentions. Instead of giving generic advice, it asks clarifying questions.

\begin{modelbox}
    To provide you with more suitable operational suggestions, I need to know some information: What is your current risk tolerance level? What is your main investment goal—capital preservation, steady growth, or aggressive returns? Also, are you looking to adjust your existing position size, or considering rebalancing across multiple funds?
\end{modelbox}

\vspace{1em}
\noindent\textbf{Step 11: Process Analysis (Round 2).} The Process Analysis confirms that the assistant's decision to ask clarifying questions is appropriate given the personalized nature of investment advice.

\begin{agentbox}[title={\faCogs\ Global Agent: Process Analysis}]
    \textbf{Assessment:}
    \begin{itemize}
        \item \textbf{Response Strategy:} Assistant chose to gather more information before providing advice. This is appropriate for personalized investment recommendations.
        \item \textbf{Question Relevance:} Questions about risk tolerance, goals, and position adjustment are directly relevant to the user's query.
        \item \textbf{Professional Standards:} Avoiding generic advice without understanding client profile aligns with responsible financial guidance.
    \end{itemize}
    
    \textbf{Result:} \framebox{True}
\end{agentbox}

\vspace{1em}
\noindent\textbf{Step 12: Dialogue Planning Update (Dynamic Insertion).} Since the assistant requested additional information, the dialogue flow has deviated from the original plan. The Global Agent detects this and dynamically updates the plan by inserting a new round for the user to respond to the clarification questions. This demonstrates the framework's ability to adapt to conversational dynamics while maintaining structural coherence.

\begin{agentbox}[title={\faClipboardList\ Global Agent: Dialogue Planning}]
    \texttt{<thought>}\\
    1. \textbf{History Locking:} Found 2 \texttt{<user>} tags. First 2 rounds are history and must be locked.\\
    2. \textbf{Latest Progress:} Assistant initiated a "Follow-up Question" in Round 2 instead of providing direct advice.\\
    3. \textbf{Update Strategy:} Triggered "Principle 2: Responsive Insertion"—when the assistant requests information, a new user round must be added.\\
    4. \textbf{New Plan:} Insert Round 3 for the user to answer the clarification questions.\\
    \texttt{</thought>}
    
    \texttt{<completion\_status>}
    
    \framebox{False}
    
    \texttt{</completion\_status>}
    
    \textbf{Updated Dialogue Plan:}
    \begin{itemize}
        \item \textbf{Round 1:} Inquire about 019667 credit risk details. \textcolor{teal}{[Completed]}
        \item \textbf{Round 2:} Follow up on countermeasures regarding low credit risk. \textcolor{teal}{[Completed—Clarification Requested]}
        \item \textbf{Round 3 (Inserted):} Provide current risk tolerance, investment goals, and position adjustment intentions.
    \end{itemize}
\end{agentbox}

\vspace{1em}
\noindent\textbf{Step 13: User Response to Clarification.} Following the updated plan, the User Agent generates a response that addresses the assistant's questions, providing information about risk tolerance, investment goals, and current sentiment.

\begin{userbox}
   I'm a pretty conservative investor—I'm mainly just trying to save up for my kid's education and my own retirement. Lately, I've been feeling a bit nervous, so I'm thinking of cutting back on my position in this fund.
\end{userbox}
The rest of the conversation flow is omitted...

\subsection{Annotation Guidelines for FinToolBench Verification}
\label{app:Annotation Guidelines for FinToolBench Verification}

To ensure the quality and reliability of FinToolBench, we employed finance-specialized annotators to verify and curate all 843 gold-standard samples. The annotation process focused on selecting high-quality instances with accurate ground-truth labels, ensuring the benchmark's validity for evaluating financial tool-calling capabilities. This section details the complete annotation protocol used to validate data quality across tool selection, parameter accuracy, and response appropriateness.

\textbf{Annotator Qualifications.} All annotators met the following criteria: (1) Bachelor's degree or higher in finance, economics, or related quantitative fields; (2) Minimum 3 years of professional experience in financial analysis, investment consulting, or financial technology; (3) Familiarity with financial APIs, data systems, and tool-based workflows; (4) Strong understanding of financial products (stocks, bonds, funds), market data structures, and regulatory requirements. Prior to annotation, all annotators underwent a calibration session using 50 pilot samples to ensure consistent interpretation of the evaluation framework.

\textbf{Annotation Task Overview.} For each sample, annotators performed three-dimensional verification to validate the correctness of reference solutions: (1) \textit{Tool Selection Correctness}: Verify whether the reference tool calls appropriately address user intent; (2) \textit{Parameter Accuracy}: Validate that all tool parameters in ground-truth labels are correctly filled with precise values and proper formatting; (3) \textit{Response Appropriateness}: Confirm that the reference response appropriately addresses the user's query. The validation criteria directly correspond to the quantitative evaluation framework defined in Section \ref{Financial Tool-Use Benchmark} (CB-HWS mechanism) and Section \ref{Evaluation Metrics} (domain-specific metrics).

\textbf{Tool Selection Validation.} Annotators applied the principles defined in Section~\ref{CBHWS}: (a) \textit{Intent Coverage}---verify that all user intents are addressed by at least one tool call; (b) \textit{Relevance}---check that each tool call directly serves a user intent (no irrelevant invocations); (c) \textit{Efficiency}---assess whether the tool combination demonstrates synergy; (d) \textit{Tool Naming}---verify exact tool name matches (case-sensitive).

\textbf{Parameter Accuracy Validation.} Annotators applied parameter validation rules defined in Section~\ref{CBHWS}, under financial domain constraints: (a) \textit{Critical Entity Precision}---stock codes must include exchange suffixes (e.g., \texttt{600519.SH}, \texttt{AAPL.O}); fund/bond codes must follow standard formats; failure triggers Circuit-Breaker ($V=0$, Equation~\ref{equation:V}). (b) \textit{Entity Consistency}---no entity hallucination; company abbreviations acceptable if unambiguous. (c) \textit{Temporal Accuracy}---dates in \texttt{YYYY-MM-DD} format; relative time expressions correctly resolved. (d) \textit{Numeric Precision}---evaluated via Key Digit Accuracy (KDA, Section~\ref{Evaluation Metrics}); unit consistency verified. (e) \textit{Schema Compliance}---all required parameters present; types match; enum constraints respected.

\textbf{Response Appropriateness Validation.} For Tool-Call instances: (a) Execution logic follows business workflows (assessed via ITA, Section~\ref{Evaluation Metrics}); (b) Tool calls collectively provide complete solutions; (c) Correct identification of parallel vs. serial execution patterns. For Non-Tool-Call instances (Table~\ref{tab:taxonomy_compact}): (a) \textit{Unavailability Detection (UD)}---no tool invoked when no suitable tool exists; (b) \textit{Clarification Inquiry (CI)}---explicitly request missing parameters; no premature tool calls; assessed via LLM Judge (Equation 6); (c) \textit{Direct Response (DR)}---provide direct answers when sufficient context available; no unnecessary tool invocations.

\textbf{Relationship to Evaluation Protocol.} The annotation guidelines described above were used to curate and validate the FinToolBench dataset. The actual evaluation of model predictions on this benchmark follows the automated scoring mechanisms detailed in Section~\ref{Financial Tool-Use Benchmark}, including the CB-HWS framework (Section~\ref{CBHWS}) and domain-specific metrics (Section~\ref{Evaluation Metrics}). Human annotators ensured the quality of ground-truth labels; model evaluation is conducted through automated quantitative assessment.

\textbf{Ethical Considerations.} All annotators participated voluntarily and were compensated at market-appropriate rates commensurate with their professional expertise and geographic location. The annotation task involved only synthetic dialogues generated by LLMs (Section~\ref{sec:FinToolSyn}); no sensitive personal information or real client data was included. Annotators were informed that their work would be used for research purposes and could withdraw at any time without penalty. The task posed minimal risk as it consisted of standard professional judgment similar to routine financial analysis work.

\section{Model Training Details}
\label{app:model_training}

\subsection{Training Procedures}
\label{app:training_procedures}

To comprehensively evaluate model performance across different interaction paradigms and scales, we conducted Supervised Fine-Tuning (SFT) on two representative base models: \textbf{Qwen3-8B} and \textbf{Qwen3-14B}. For each base model, we implemented two distinct fine-tuning strategies, resulting in a total of \textbf{four fine-tuned variants}:

\begin{itemize}
    \item \textbf{Text-based Prompting (Prompt Mode)}: Tool definitions and descriptions are embedded directly into the system prompt. The model is trained to generate tool invocation requests in a specific text format.
    \item \textbf{Native Function Calling (FC Mode)}: The model is optimized to leverage inherent tool-use capabilities, generating structured call instructions directly compatible with standard API definitions.
\end{itemize}

All fine-tuning experiments were conducted on a cluster of \textbf{8 NVIDIA H20 GPUs}. We utilized the PyTorch framework with \textbf{LoRA (Low-Rank Adaptation)} for parameter-efficient fine-tuning. To ensure fair comparison, we maintained consistent hyperparameters across all four training runs. We targeted all linear modules (e.g., $q\_proj$, $k\_proj$, $v\_proj$) and trained for 1 epoch using a global batch size of 128 (achieved via gradient accumulation). Detailed hyperparameters are provided in Table~\ref{tab:hyperparameters}.

\begin{table}[h]
    \centering
    \small
    \begin{tabular}{lc}
        \toprule
        \textbf{Hyperparameter} & \textbf{Value} \\
        \midrule
        Optimizer & AdamW (Fused) \\
        Learning Rate & $2 \times 10^{-4}$ \\
        LR Scheduler & Cosine \\
        Warmup Ratio & 0.1 \\
        Global Batch Size & 128 \\
        Gradient Accumulation Steps & 16 \\
        Epochs & 1 \\
        Max Sequence Length & 32,768 \\
        Weight Decay & 0.01 \\
        BF16 Precision & True \\
        \midrule
        \textit{LoRA Configuration} & \\
        LoRA Rank ($r$) & 16 \\
        LoRA Alpha ($\alpha$) & 32 \\
        LoRA Dropout & 0.05 \\
        Target Modules & All Linear \\
        \bottomrule
    \end{tabular}
    \caption{Hyperparameters for Supervised Fine-Tuning (Applied to all 8B/14B and Prompt/FC models).}
    \label{tab:hyperparameters}
\vspace{-4mm}
\end{table}

\subsection{Training Data Processing}
\label{app:data_processing}

The raw synthetic data generated in our pipeline adheres to the \textbf{Model Context Protocol (MCP)}, which defines tool interfaces using a specific schema structure. However, to align with the standard training paradigms of most open-source Large Language Models (LLMs), we performed a format standardization process.

\paragraph{Standardization for Native FC Mode}
For the Native Function Calling (FC) experiments, we converted the MCP tool definitions into the standard function-calling format (compatible with OpenAI/Qwen JSON Schema). This involves mapping MCP's \texttt{inputSchema} to the standard \texttt{parameters} field and restructuring the conversation history to utilize the model's native special tokens (e.g., \texttt{<tool\_call>}). Figure~\ref{fig:mcp_to_fc} illustrates a comparison between the original MCP format and the converted standard format.

\begin{figure}[h]
\begin{lstlisting}[basicstyle=\ttfamily\scriptsize, frame=single, breaklines=true, captionpos=b]
# 1. Original MCP Format
{
  "name": "get_stock_price",
  "inputSchema": {
    "type": "object",
    "properties": {
      "symbol": {"type": "string"}
    },
    "required": ["symbol"]
  }
}

# ----------------

# 2. Converted Input Schema (FC Mode)
{
  "type": "function",
  "function": {
    "name": "get_stock_price",
    "description": "Get the current stock price...",
    "parameters": {
      "type": "object",
      "properties": {
        "symbol": {"type": "string"}
      },
      "required": ["symbol"]
    }
  }
}

# ----------------

# 3. Expected Model Output (FC Mode)
<tool_call>
{"name": "get_stock_price", "arguments": {"symbol": "600519.SH"}}
</tool_call>
\end{lstlisting}
\caption{FC Mode data pipeline: (1) original MCP tool schema, (2) converted input schema passed to the model API, and (3) expected model output format verified by the Format Compliance check in CB-HWS Phase~1.}
\vspace{-4mm}
\label{fig:mcp_to_fc}
\end{figure}

\paragraph{System Prompt for Prompt Mode}
For the Text-based Prompting experiments, we did not use special tokens. Instead, we flattened the tool definitions into a text block and introduced a comprehensive \textbf{System Prompt} (Figure~\ref{fig:system_prompt_for_prompt_mode}) to guide the model's behavior. This prompt explicitly instructs the model on how to format function calls and handle edge cases (e.g., missing parameters). The specific system prompt used is as follows:
\begin{figure*}
\begin{promptbox}[System Prompt for Prompt Mode]
You are an expert in composing functions. You are given a question and a set of possible functions. Based on the question, you will need to make one or more function/tool calls to achieve the purpose. If none of the functions can be used, point it out. If the given question lacks the parameters required by the function, also point it out.

You should only return the function calls in your response.

If you decide to invoke any of the function(s), you MUST put it in the format of \$\{escaped\_format\_instruction\}. You SHOULD NOT include any other text in the response.

At each turn, you should try your best to complete the tasks requested by the user within the current turn. Continue to output functions to call until you have fulfilled the user's request to the best of your ability. Once you have no more functions to call, the system will consider the current turn complete and proceed to the next turn or task.

Here is a list of functions in json format that you can invoke.
\$\{functions\}
\end{promptbox}
\caption{System Prompt for Prompt Mode}
\label{fig:system_prompt_for_prompt_mode}
\vspace{-4mm}
\end{figure*}

\subsection{Data Sampling Strategy}
\label{app:sampling_strategy}

To ensure the model possesses robust generalization capabilities across varying degrees of retrieval noise and functional complexity, we implemented a \textbf{two-stage stratified sampling strategy} to construct the final dataset of \textbf{100,000 instances}.

\paragraph{Stage 1: Retrieval Environment Stratification}
First, to prevent the model from overfitting to a specific candidate tool distribution, we partitioned the training data equally across the three retrieval configurations defined in Section~\ref{subsec:Globally-Controlled Dynamic Retrieval Dialogue Generation} (approx. 33.3\% each):
\begin{itemize}
    \item \textbf{Static Retrieval}: Simulates stable, closed-domain scenarios with fixed candidate sets.
    \item \textbf{Vector Retrieval}: Introduces semantic relevance noise, requiring the model to identify tools based on embedding similarity.
    \item \textbf{Graph-Enhanced Retrieval}: Represents the most challenging "noisy" environment, where the candidate set includes hard negatives (e.g., semantically related but functionally distinct tools) introduced by the tool dependency graph.
\end{itemize}

\paragraph{Stage 2: Functional Category Stratification}
Second, within each retrieval subset, we applied a fine-grained stratification based on the taxonomy presented in \textbf{Table 2}. This ensures a balanced coverage of interaction logic:
\begin{itemize}
    \item \textbf{Tool-Call Scenarios}: Covering \textbf{Single-tool} vs. \textbf{Multi-tool} contexts and diverse execution patterns (\textbf{Single}, \textbf{Parallel}, \textbf{Serial}), forcing the model to handle both simple execution and complex dependency planning.
    \item \textbf{Non-Tool-Call Scenarios}: Explicitly including \textbf{Unavailability Detection (UD)}, \textbf{Clarification Inquiry (CI)}, and \textbf{Direct Response (DR)} to train the model's rejection and clarification capabilities when suitable tools are absent or context is insufficient.
\end{itemize}

By combining these two stages, the final corpus $D_{train}$ forces the model to learn robust decision boundaries: identifying the correct tool (or deciding not to call one) regardless of whether the retrieved candidate set is clean (Static) or noisy (Graph-Enhanced).

\subsection{Data Quality Control}
\label{app:data_quality}

\subsubsection{Automated Filtering Pipeline}

Throughout the synthesis process, the Global Agent enforces strict quality control at two granularities.

\textbf{Tool-Level Executability.} Each synthesized tool undergoes syntax validation (parsable JSON schema) and strict API execution checks: endpoints must return valid schemas; null returns or HTTP errors trigger a rejection. The LLM is given up to $T_{\max}=3$ attempts to self-correct hallucinated or ill-formed tool definitions.

\textbf{Dialogue-Level Logic.} The Global Agent verifies that (1) extracted parameters match the API schema, (2) the assistant's final response is faithfully grounded in the Tool Agent's execution output (no hallucinated information), and (3) the dialogue trajectory is logically consistent with the evolving dialogue plan.

\subsubsection{Discard Rate and Failure Analysis}

The overall discard rate during the synthesis of the 148k-instance corpus was strictly controlled below \textbf{10\%}. Discarded samples primarily fail due to three causes (in descending frequency):

\begin{enumerate}
    \item \textbf{Entity/Ticker Hallucination} (most common): The LLM hallucinates an incorrect stock identifier format. For example, when asked about ``Kweichow Moutai'', the model may output \texttt{ticker="Moutai"} instead of the required \texttt{"600519.SH"}. The API execution fails, and if the model cannot self-correct within $T_{\max}$ retries, the trajectory is discarded.
    \item \textbf{Empty Execution Returns}: The API executes successfully but returns empty data (e.g., querying a financial report on a date it was not published), yielding an uninformative dialogue that cannot serve as a training example.
    \item \textbf{Reflection Loops}: The Global Agent detects a logical inconsistency but the generation agent fails to produce a valid correction within the retry budget.
\end{enumerate}

\subsubsection{Human Verification}

To validate the overall quality of the synthesized data, we employed finance-specialized annotators (3+ years of professional experience) to perform multi-dimensional verification on a randomly sampled subset. Annotators evaluated Tool Selection Correctness, Parameter Accuracy (including critical entity precision, temporal accuracy, and numeric precision), and Response Appropriateness. Against these strict financial-domain criteria, the sampled subset achieved a \textbf{human acceptance rate of 94.2\%}, confirming the pipeline's reliability.

\section{Evaluation Details}
\label{app:evaluation}

\subsection{Main Experiment Results}
\label{app:total_results}

To rigorously evaluate the planning and execution capabilities of LLMs within complex financial scenarios, we present a granular breakdown of performance in Table 3. We introduce a taxonomy of tool-use complexity based on the execution logic required to resolve user intents:
\begin{itemize}
    \item Single-Tool (single): Atomic execution where the user's intent is resolved by a single, independent tool invocation. This setting primarily assesses basic semantic alignment and parameter extraction.
    \item Parallel Execution (parallel): Scenarios requiring the simultaneous invocation of multiple tools within a single turn. This tests the model's ability to decompose composite queries into independent sub-tasks (e.g., retrieving stock prices for multiple companies concurrently) without hallucinating dependencies.
    \item Serial Execution (serial): The most cognitively demanding scenario, involving a dependency chain spanning multiple turns. The model must execute a tool, interpret the intermediate output, and utilize this context to formulate subsequent tool calls (e.g., retrieving a stock code to subsequently query a financial report).
\end{itemize}

\paragraph{Overall Performance and Robustness}
As shown in Table~\ref{tab:tool-call-total}, our fine-tuned models demonstrate state-of-the-art performance across both interaction paradigms, effectively bridging the gap between open-source models and proprietary frontiers.
In Prompt Mode (Panel A), FinTool-Qwen3-14B achieves the highest average accuracy of 71.06, significantly surpassing strong open-source baselines like DeepSeek-V3.1-Terminus (68.27) and proprietary models like GPT-4o (62.31).
In Function Calling (FC) Mode (Panel B), the efficiency of our method is even more evident. FinTool-Qwen3-14B achieves an impressive average of 70.03, outperforming the massive Qwen3-235B-Instruct (67.00) and GPT-4o (63.86). This result suggests that domain-specific fine-tuning with Forward Synthesis data can enable smaller models to surpass significantly larger general-purpose models in specialized tool-use tasks.

\paragraph{Mastery of Complex Temporal Dependencies}
Performance generally exhibits a downward trend as complexity increases (single $\rightarrow$ parallel $\rightarrow$ serial). However, our models exhibit exceptional resilience in Serial Execution, which serves as the upper bound of difficulty in our benchmark.
\begin{itemize}
    \item Prompt Mode: In the Multi-Turn Multi-Tool Serial setting, FinTool-Qwen3-14B achieves a score of 61.62. In stark contrast, DeepSeek-V3.1-Terminus, despite its strong general reasoning capabilities, drops to 30.83, and GPT-4o achieves 46.69. This indicates that general-purpose instruction tuning often fails to capture the long-horizon state dependencies required for financial analysis, whereas our method instills this specific reasoning capability.
    \item FC Mode: The advantage persists in native function calling. FinTool-Qwen3-14B scores 57.03 in the serial setting, surpassing both Qwen3-235B (54.97) and GPT-4o (47.25). Meanwhile, base models like Qwen2.5-72B collapse to 14.47. This confirms that our model does not merely memorize API schemas but learns the underlying logic of sequential problem-solving.
\end{itemize}

\paragraph{Non-Tool-Call Capabilities}
A reliable agent must not only invoke tools correctly but also discern when not to use them. Table~\ref{tab:non-tool-call-total} presents the performance in non-tool scenarios.
Some baselines, such as Gemini-2.5-Flash, achieve near 100\% scores in "No-Tool-Reply" but fail to execute actual tools (11.86\% Overall Acc in Table 3), indicating a tendency towards over-refusal. Conversely, FinTool-Qwen3-14B strikes a critical balance. It maintains high accuracy in Normal-Reply (74.07 in Prompt Mode) and Follow-up questions, proving that the model has learned to distinguish between executable financial tasks, general chit-chat, and ambiguous queries requiring clarification, rather than simply overfitting to tool invocation triggers.

\subsection{CB-HWS Algorithm Details}
\label{app:cbhws_algo}

The following pseudocode details the CB-HWS mechanism described in Section~\ref{CBHWS}. Algorithm~\ref{alg:cb_hws_main} presents the main control flow, and Algorithm~\ref{alg:tool_call_scoring} details the tool-call circuit-breaker scoring logic.

\begin{algorithm*}[!ht]
\SetKwInput{Input}{Input}
\SetKwInput{Output}{Output}
\SetKwFunction{CalcToolScore}{CalcToolCallScore}
\caption{CB-HWS Main Control Flow}
\label{alg:cb_hws_main}
\Input{Pred.\ $\mathcal{T}_{\text{pred}}$, gold $\mathcal{T}_{\text{gold}}$, context $\mathcal{C}$}
\Output{Total score $S_{\text{total}}$}
\BlankLine
\eIf{instance is a \textbf{Tool-Call} case}{
    Init $L \leftarrow [\,]$, turn count $T$\;
    \ForEach{turn $t \in [1, T]$}{
        $S^{(t)} \leftarrow$ \CalcToolScore{$\mathcal{T}_{\text{pred}}^{(t)}$, $\mathcal{T}_{\text{gold}}^{(t)}$, $\mathcal{C}$}\;
        Append $S^{(t)}$ to $L$\;
    }
    $S_{\text{total}} \leftarrow \frac{1}{T}\sum S^{(t)}$\;
}{
    \tcc{Non-Tool-Call case}
    \tcc{Phase 1: Anti-Hallucination check}
    \If{any tool call in $\mathcal{T}_{\text{pred}}$}{
        \Return $0$\;
    }
    \tcc{Phase 2: Intent verification}
    $\delta \leftarrow 1$\;
    \If{\textbf{CI} (Clarification Inquiry)}{
        $\delta \leftarrow \mathbb{I}\bigl(\textsc{LLM\_Judge}(\mathcal{T}_{\text{pred}})\bigr)$\;
    }
    $S_{\text{total}} \leftarrow \delta \times 100$\;
}
\Return $S_{\text{total}}$\;
\end{algorithm*}

\begin{algorithm*}[!ht]
\SetKwInput{Input}{Input}
\SetKwInput{Output}{Output}
\SetKwInput{Hyperparams}{Hyperparams}
\caption{Tool-Call CB Scoring ($S_{\text{TC}}$)}
\label{alg:tool_call_scoring}
\Input{Pred.\ $\mathcal{T}_{\text{pred}}$, gold $\mathcal{T}_{\text{gold}}$, context $\mathcal{C}$}
\Output{Turn score $S_{\text{turn}} \in [0,100]$}
\Hyperparams{$w_s\!=\!0.3,\,w_v\!=\!0.7,\,w_k\!=\!0.4,\,w_e\!=\!0.6$}
\BlankLine
\tcc{Phase 1: Rule-Based Circuit Breaker}
\If{fmt.\ error \textbf{or} hallucination \textbf{or} schema violation}{
    \Return $0$\;
}
\BlankLine
\tcc{Phase 2: LLM-Based Soft Evaluation}
\tcc{Step 1: Tool selection score}
$k \leftarrow \textsc{Judge}_{\text{sel}}(\mathcal{T}_{\text{pred}},\mathcal{T}_{\text{gold}})$\;
\If{$k = 0$}{
    \Return $0$\;
}
\tcc{Step 2: Param execution scoring}
$S_e \leftarrow 0$,\; $N \leftarrow |\mathcal{T}_{\text{pred}}|$\;
\ForEach{tool call $i$ in $\mathcal{T}_{\text{pred}}$}{
    $x_i \leftarrow \textsc{Judge}_{\text{name}}(i)$\;
    $\{y_{ij}\} \leftarrow \textsc{Judge}_{\text{val}}(i)$\;
    \eIf{$m_i > 0$}{
        $s_i \leftarrow w_s \cdot x_i + w_v \cdot \frac{1}{m_i}\sum_j y_{ij}$\;
    }{
        $s_i \leftarrow x_i$\;
    }
    $S_e \leftarrow S_e + s_i$\;
}
$S_e \leftarrow S_e / N$\;
\tcc{Phase 3: Hierarchical aggregation}
$S_{\text{turn}} \leftarrow (w_k \cdot k + w_e \cdot S_e) \times 10$\;
\Return $S_{\text{turn}}$\;
\end{algorithm*}

\begin{table*}[htbp]
  \centering
  \small
  \resizebox{\textwidth}{!}{%
    \begin{tabular}{lccccccccccccccc}
      \toprule
      \multirow{4}{*}{\textbf{Model}} & \multirow{4}{*}{\textbf{Avg}} & \multicolumn{7}{c}{\textbf{Single-Turn}} & \multicolumn{7}{c}{\textbf{Multi-Turn}} \\
      \cmidrule(lr){3-9} \cmidrule(lr){10-16}
      & & \multicolumn{3}{c}{\textbf{Single-Tool}} & \multicolumn{4}{c}{\textbf{Multi-Tool}} & \multicolumn{3}{c}{\textbf{Single-Tool}} & \multicolumn{4}{c}{\textbf{Multi-Tool}} \\
      \cmidrule(lr){3-5} \cmidrule(lr){6-9} \cmidrule(lr){10-12} \cmidrule(lr){13-16}
      & & \multirow{2}{*}{\textbf{Avg}} & \multirow{2}{*}{\textbf{single}} & \multirow{2}{*}{\textbf{parallel}} & \multirow{2}{*}{\textbf{Avg}} & \multirow{2}{*}{\textbf{single}} & \multirow{2}{*}{\textbf{parallel}} & \multirow{2}{*}{\textbf{serial}} & \multirow{2}{*}{\textbf{Avg}} & \multirow{2}{*}{\textbf{single}} & \multirow{2}{*}{\textbf{parallel}} & \multirow{2}{*}{\textbf{Avg}} & \multirow{2}{*}{\textbf{single}} & \multirow{2}{*}{\textbf{parallel}} & \multirow{2}{*}{\textbf{serial}} \\
      & & & & & & & & & & & & & & & \\
      \midrule
      
      \multicolumn{16}{c}{\cellcolor{gray!10}\textbf{Panel A: Prompt Mode}} \\
      \midrule
      \multicolumn{16}{r}{\textit{Closed-Source Models}} \\
      Claude-4.0-Sonnet & 44.19 & 58.09 & 73.93 & 42.25 & 37.59 & 45.81 & 25.98 & 40.99 & 37.09 & 37.18 & 36.99 & 43.98 & 52.20 & 39.46 & 40.28 \\
      Gemini-2.5-Flash & 11.86 & 1.92 & 0.00 & 3.84 & 4.60 & 1.76 & 1.96 & 10.08 & 22.22 & 12.89 & 31.56 & 18.71 & 13.72 & 23.74 & 18.67 \\
      Gemini-2.5-Pro & 72.12 & 81.56 & 83.90 & 79.21 & 72.34 & 81.19 & 74.85 & 60.98 & 67.32 & 62.93 & 71.71 & 67.27 & 71.65 & 68.25 & 61.90 \\
      GPT-3.5-Turbo & 44.07 & 56.77 & 76.12 & 37.41 & 39.48 & 40.80 & 35.95 & 41.70 & 42.75 & 50.98 & 34.53 & 37.26 & 39.69 & 48.35 & 23.72 \\
      GPT-4o & 62.31 & 74.87 & 71.36 & 78.38 & 58.02 & 79.37 & 43.93 & 50.75 & 62.36 & 59.53 & 65.19 & 53.98 & 64.11 & 51.14 & 46.69 \\
      \midrule
      \multicolumn{16}{r}{\textit{Open-Source Models}} \\
      DeepSeek-V3.1-Terminus & 68.27 & 79.83 & 77.78 & 81.87 & 56.83 & 75.04 & 60.22 & 35.24 & 73.61 & 71.62 & 75.60 & 62.82 & 80.58 & 77.04 & 30.83 \\
      Qwen3-235B-Instruct & 67.65 & 77.19 & 80.90 & 73.49 & 59.11 & 63.93 & 56.79 & 56.61 & 67.79 & 72.69 & 62.89 & 66.50 & 69.41 & 70.50 & 59.60 \\
      Qwen2.5-72B-Instruct & 66.36 & 78.12 & 78.73 & 77.50 & 59.69 & 65.78 & 49.68 & 63.62 & 64.79 & 70.21 & 59.36 & 62.84 & 70.46 & 66.94 & 51.11 \\
      Qwen2.5-32B-Instruct & 54.10 & 66.26 & 71.91 & 60.61 & 45.39 & 60.89 & 33.72 & 41.56 & 53.43 & 60.46 & 46.40 & 51.33 & 62.14 & 53.95 & 37.89 \\
      Qwen2.5-14B-Instruct & 56.84 & 68.47 & 70.14 & 66.80 & 52.06 & 62.19 & 41.84 & 52.14 & 54.91 & 61.77 & 48.05 & 51.93 & 65.21 & 58.07 & 32.51 \\
      Qwen2.5-7B-Instruct & 53.58 & 61.10 & 68.89 & 53.32 & 45.12 & 44.24 & 43.00 & 48.11 & 60.17 & 69.44 & 50.90 & 47.93 & 48.26 & 49.28 & 46.24 \\
      Qwen3-32B & 66.52 & 79.82 & 78.52 & 81.11 & 57.85 & 65.79 & 51.55 & 56.20 & 63.60 & 72.02 & 55.17 & 64.79 & 72.30 & 72.53 & 49.53 \\
      Qwen3-14B & 60.60 & 67.43 & 71.68 & 63.19 & 51.69 & 58.93 & 39.81 & 56.32 & 64.68 & 69.59 & 59.77 & 58.58 & 67.83 & 61.01 & 46.91 \\
      Qwen3-8B & 57.60 & 70.24 & 74.04 & 66.44 & 47.04 & 49.92 & 37.98 & 53.21 & 59.42 & 66.54 & 52.30 & 53.71 & 60.96 & 61.51 & 38.65 \\
      \midrule
      \multicolumn{16}{r}{\textit{FinTool-Tuned Models (Ours)}} \\
      FinTool-Qwen3-14B & \textbf{71.06} & 77.23 & 76.39 & 78.07 & 58.24 & 55.07 & 59.53 & 60.11 & 79.99 & 80.13 & 79.84 & 68.78 & 75.19 & 69.52 & 61.62 \\
      FinTool-Qwen3-8B & 70.85 & 78.10 & 75.89 & 80.32 & 58.12 & 60.17 & 57.23 & 56.97 & 79.37 & 85.24 & 73.50 & 67.82 & 75.12 & 66.20 & 62.14 \\
      
      \midrule
      \multicolumn{16}{c}{\cellcolor{gray!10}\textbf{Panel B: Function Calling Mode}} \\
      \midrule
      \multicolumn{16}{r}{\textit{Closed-Source Models}} \\
      Claude-4.0-Sonnet & 65.27 & 67.28 & 72.18 & 62.38 & 62.15 & 70.94 & 61.01 & 54.50 & 68.27 & 70.13 & 66.41 & 63.36 & 66.14 & 63.20 & 60.76 \\
      Gemini-2.5-Flash & 41.96 & 57.73 & 65.10 & 50.35 & 28.50 & 37.71 & 24.87 & 22.91 & 42.80 & 44.25 & 41.35 & 38.80 & 44.91 & 45.86 & 25.62 \\
      Gemini-2.5-Pro & 53.30 & 64.77 & 66.11 & 63.42 & 47.68 & 48.85 & 58.43 & 35.77 & 49.56 & 54.97 & 44.14 & 51.17 & 56.83 & 55.47 & 41.21 \\
      GPT-3.5-Turbo & 51.20 & 68.51 & 66.20 & 70.81 & 37.15 & 42.35 & 34.44 & 34.67 & 53.36 & 58.89 & 47.84 & 45.78 & 50.30 & 52.93 & 34.11 \\
      GPT-4o & 63.86 & 74.53 & 75.65 & 73.40 & 53.25 & 63.93 & 53.87 & 41.96 & 71.56 & 67.27 & 75.84 & 56.10 & 56.90 & 64.15 & 47.25 \\
      \midrule
      \multicolumn{16}{r}{\textit{Open-Source Models}} \\
      Qwen3-235B-Instruct & 67.00 & 76.98 & 79.98 & 73.98 & 59.83 & 65.19 & 61.96 & 52.34 & 64.04 & 65.84 & 62.24 & 67.14 & 77.76 & 68.69 & 54.97 \\
      Qwen2.5-72B-Instruct & 14.62 & 4.29 & 4.38 & 4.21 & 6.83 & 9.98 & 6.59 & 3.93 & 18.57 & 15.65 & 21.49 & 28.80 & 35.70 & 36.23 & 14.47 \\
      Qwen2.5-32B-Instruct & 43.44 & 62.77 & 67.81 & 57.74 & 44.10 & 53.06 & 46.92 & 32.30 & 33.95 & 31.98 & 35.91 & 32.92 & 40.65 & 36.85 & 21.24 \\
      Qwen2.5-14B-Instruct & 45.55 & 69.92 & 73.37 & 66.46 & 41.10 & 49.21 & 40.38 & 33.73 & 38.67 & 40.56 & 36.79 & 32.52 & 37.52 & 29.12 & 30.93 \\
      Qwen2.5-7B-Instruct & 44.77 & 54.40 & 59.39 & 49.42 & 43.49 & 59.68 & 37.81 & 32.98 & 42.60 & 52.01 & 33.19 & 38.60 & 52.09 & 41.08 & 22.62 \\
      Qwen3-32B & 55.75 & 73.13 & 71.02 & 75.24 & 50.14 & 60.68 & 51.05 & 38.70 & 48.23 & 56.83 & 39.63 & 51.49 & 65.32 & 54.58 & 34.55 \\
      Qwen3-14B & 51.85 & 68.25 & 74.15 & 62.35 & 51.13 & 59.82 & 49.44 & 44.12 & 43.64 & 49.55 & 37.74 & 44.39 & 57.92 & 49.45 & 25.82 \\
      Qwen3-8B & 48.03 & 66.55 & 69.66 & 63.45 & 43.37 & 51.47 & 45.59 & 33.06 & 39.22 & 38.12 & 40.32 & 42.97 & 57.20 & 47.54 & 24.16 \\
      \midrule
      \multicolumn{16}{r}{\textit{FinTool-Tuned Models (Ours)}} \\
      FinTool-Qwen3-14B & \textbf{70.03} & 77.86 & 80.80 & 74.91 & 59.11 & 62.34 & 53.51 & 61.49 & 76.98 & 82.05 & 71.92 & 66.18 & 74.05 & 67.47 & 57.03 \\
      FinTool-Qwen3-8B & 68.31 & 77.20 & 73.88 & 80.53 & 56.01 & 56.31 & 51.40 & 60.33 & 74.53 & 75.66 & 73.40 & 65.50 & 74.97 & 66.11 & 55.43 \\
      
      \bottomrule
    \end{tabular}
  }
  \caption{Detailed Tool-Call Performance Breakdown. Overall Acc is calculated as the average of OA across four scenarios.}
  \label{tab:tool-call-total}
\end{table*}

\begin{table*}[htbp]
    \centering
    \resizebox{\textwidth}{!}{
      \begin{tabular}{lcccccccccc}
        \toprule
      \multicolumn{11}{c}{\cellcolor{gray!10}\textbf{Panel A: Prompt Mode}} \\
        \midrule
        \multirow{2}{*}{\textbf{Model}} & \multirow{2}{*}{\textbf{Overall Acc}} & \multicolumn{3}{c}{\textbf{No-Tool-Reply}} & \multicolumn{3}{c}{\textbf{Follow-up}} & \multicolumn{3}{c}{\textbf{Normal-Reply}} \\
        \cmidrule(lr){3-5} \cmidrule(lr){6-8} \cmidrule(lr){9-11}
        & & \textbf{Overall} & \textbf{Single-Turn} & \textbf{Multi-Turn} & \textbf{Overall} & \textbf{Single-Turn} & \textbf{Multi-Turn} & \textbf{Overall} & \textbf{Single-Turn} & \textbf{Multi-Turn} \\
        \midrule
        \multicolumn{11}{r}{\textit{Closed-Source Models}} \\
        Claude-4.0-Sonnet & 57.96 & 100.00 & 31.73 & 30.77 & 38.46 & 26.83 & 50.00 & 97.25 & 27.59 & 78.57 \\
        Gemini-2.5-Flash & 40.85 & 100.00 & 38.46 & 36.54 & 25.93 & 10.25 & 100.00 & 100.00 & 25.93 & 10.25 \\
        Gemini-2.5-Pro & 62.15 & 54.64 & 25.00 & 21.15 & 66.90 & 60.71 & 73.08 & 14.81 & 72.75 & 54.64 \\
        GPT-3.5-Turbo & 38.47 & 41.72 & 4.81 & 0.00 & 46.50 & 16.07 & 76.92 & 11.11 & 45.04 & 41.72 \\
        GPT-4o & 59.67 & 65.47 & 28.85 & 21.15 & 74.18 & 71.43 & 76.92 & 37.04 & 61.99 & 65.47 \\
        \midrule
        \multicolumn{11}{r}{\textit{Open-Source Models}} \\
        DeepSeek-V3.1-Terminus & 55.32 & 47.81 & 39.62 & 56.00 & 62.50 & 50.00 & 75.00 & 3.70 & 67.04 & 47.81 \\
        Qwen3-235B-Instruct & 53.05 & 43.58 & 1.92 & 0.00 & 55.22 & 64.29 & 46.15 & 0.00 & 67.02 & 43.58 \\
        Qwen2.5-72B-Instruct & 51.86 & 36.15 & 5.77 & 0.00 & 55.63 & 53.57 & 57.69 & 11.11 & 66.92 & 36.15 \\
        Qwen2.5-32B-Instruct & 51.18 & 65.08 & 3.85 & 0.00 & 72.94 & 78.57 & 67.31 & 3.70 & 53.81 & 65.08 \\
        Qwen2.5-14B-Instruct & 55.75 & 76.74 & 6.73 & 3.85 & 79.40 & 85.71 & 73.08 & 3.70 & 56.99 & 76.74 \\
        Qwen2.5-7B-Instruct & 41.67 & 30.81 & 0.00 & 0.00 & 46.57 & 64.29 & 28.85 & 0.00 & 51.67 & 30.81 \\
        Qwen3-32B & 51.67 & 36.92 & 3.85 & 0.00 & 54.88 & 48.21 & 61.54 & 0.00 & 66.20 & 36.92 \\
        Qwen3-14B & 47.87 & 36.21 & 1.92 & 1.92 & 54.60 & 55.36 & 53.85 & 0.00 & 58.59 & 36.21 \\
        Qwen3-8B & 47.55 & 46.09 & 1.92 & 1.92 & 54.46 & 58.93 & 50.00 & 3.70 & 57.37 & 46.09 \\
        \midrule
        \multicolumn{11}{r}{\textit{FinTool-Tuned Models (Ours)}} \\
        FinTool-Qwen3-14B & 57.95 & 30.32 & 50.00 & 46.15 & 41.07 & 32.14 & 50.00 & 74.07 & 68.82 & 30.32 \\
        FinTool-Qwen3-8B & 55.68 & 30.32 & 42.31 & 40.38 & 33.72 & 23.21 & 44.23 & 70.37 & 67.79 & 30.32 \\
        \bottomrule
      \multicolumn{11}{c}{\cellcolor{gray!10}\textbf{Panel B: Function Calling Mode}} \\
        \midrule
        \multirow{2}{*}{\textbf{Model}} & \multirow{2}{*}{\textbf{Overall Acc}} & \multicolumn{3}{c}{\textbf{No-Tool-Reply}} & \multicolumn{3}{c}{\textbf{Follow-up}} & \multicolumn{3}{c}{\textbf{Normal-Reply}} \\
        \cmidrule(lr){3-5} \cmidrule(lr){6-8} \cmidrule(lr){9-11}
        & & \textbf{Overall} & \textbf{Single-Turn} & \textbf{Multi-Turn} & \textbf{Overall} & \textbf{Single-Turn} & \textbf{Multi-Turn} & \textbf{Overall} & \textbf{Single-Turn} & \textbf{Multi-Turn} \\
        \midrule
        \multicolumn{11}{r}{\textit{Closed-Source Models}} \\
        Claude-4.0-Sonnet & 55.93 & 44.70 & 32.65 & 26.83 & 53.08 & 27.59 & 78.57 & 44.44 & 65.33 & 44.70 \\
        Gemini-2.5-Flash & 58.35 & 94.27 & 50.37 & 65.96 & 96.00 & 92.00 & 100.00 & 43.48 & 41.66 & 94.27 \\
        Gemini-2.5-Pro & 60.98 & 85.07 & 40.75 & 51.06 & 87.83 & 84.00 & 91.67 & 34.78 & 53.94 & 85.07 \\
        GPT-3.5-Turbo & 43.67 & 44.78 & 4.17 & 0.00 & 51.91 & 10.64 & 93.18 & 0.00 & 48.94 & 44.78 \\
        GPT-4o & 58.26 & 63.01 & 6.25 & 0.00 & 83.13 & 79.17 & 87.10 & 11.11 & 62.59 & 63.01 \\
        \midrule
        \multicolumn{11}{r}{\textit{Open-Source Models}} \\
        Qwen3-235B-Instruct & 58.06 & 59.47 & 9.62 & 5.77 & 69.37 & 71.43 & 67.31 & 3.70 & 66.61 & 59.47 \\
        Qwen2.5-72B-Instruct & 53.23 & 38.23 & 6.12 & 0.00 & 57.34 & 53.57 & 61.54 & 11.50 & 67.10 & 38.23 \\
        Qwen2.5-32B-Instruct & 52.85 & 67.34 & 4.12 & 0.00 & 74.56 & 78.57 & 70.83 & 4.10 & 54.20 & 67.34 \\
        Qwen2.5-14B-Instruct & 57.23 & 78.34 & 7.12 & 0.00 & 81.23 & 85.71 & 76.92 & 4.10 & 57.50 & 78.34 \\
        Qwen2.5-7B-Instruct & 43.23 & 32.34 & 1.12 & 0.00 & 48.23 & 53.57 & 42.86 & 1.10 & 52.10 & 32.34 \\
        Qwen3-32B & 49.58 & 53.09 & 3.85 & 0.00 & 67.10 & 55.36 & 78.85 & 0.00 & 54.71 & 53.09 \\
        Qwen3-14B & 50.47 & 63.75 & 4.81 & 0.00 & 77.34 & 64.29 & 90.38 & 0.00 & 53.13 & 63.75 \\
        Qwen3-8B & 50.05 & 73.42 & 3.85 & 1.92 & 80.98 & 69.64 & 92.31 & 3.70 & 49.05 & 73.42 \\
        \midrule
        \multicolumn{11}{r}{\textit{FinTool-Tuned Models (Ours)}} \\
        FinTool-Qwen3-14B & 57.77 & 33.32 & 45.19 & 38.46 & 45.74 & 35.71 & 55.77 & 74.07 & 68.17 & 33.32 \\
        FinTool-Qwen3-8B & 57.64 & 42.98 & 44.23 & 38.46 & 42.99 & 32.14 & 53.85 & 59.26 & 66.26 & 42.98 \\
        \bottomrule
    \end{tabular}
    }
    \caption{Non-Tool-Call Performance Comparison}
    \label{tab:non-tool-call-total}
\end{table*}

\subsection{Ablation Experiment Results}
To validate the effectiveness of our proposed synthesis components, we conducted a comprehensive ablation study on the Qwen3-8B model in Function Calling (FC) mode. We compare the following variants:
\begin{enumerate}
    \item Base (Qwen3-8B-FC): The original model without fine-tuning.
    \item Reverse Synthesis (FinTool-Ablation-reverse): Fine-tuned on data generated via the traditional "Tool $\rightarrow$ Query" paradigm.
    \item Static Retrieval (FinTool-Ablation-static): Fine-tuned on Forward Synthesis data but with fixed, clean candidate tool sets (no retrieval noise).
    \item Full Method (FinTool-Qwen3-8B-FC): The complete FinToolSyn framework with Forward Synthesis and Dynamic Retrieval.
\end{enumerate}

\paragraph{Impact on Tool Execution}
Table~\ref{tab:ablation-tool-call-total} presents the detailed breakdown of tool-calling capabilities.

The "Artificial Explicitness" Trap: The Reverse Synthesis model achieves high scores in simple tool-use metrics (e.g., 73.85 in ST-ST-single), sometimes even surpassing the Full Method. However, this comes at a cost. As shown in the Non-Tool analysis below, this model overfits to tool invocation, losing the ability to handle ambiguity.
Robustness via Dynamic Retrieval: Comparing Static vs. Full, the Full Method (trained with dynamic retrieval noise) demonstrates superior robustness in complex scenarios. In the challenging Multi-Turn Multi-Tool Serial setting, the Full Method scores 55.43, outperforming the Static model (54.02). This confirms that exposing the model to noisy candidate sets during training improves its ability to filter irrelevant tools and maintain logical chains.

\paragraph{Impact on Non-Tool Capabilities}
Table~\ref{tab:ablation-non-tool-call-total} highlights the critical role of Forward Synthesis in maintaining a balanced agent.

Rejection and Clarification: The Reverse Synthesis model suffers a catastrophic collapse in non-tool capabilities. Its No-Tool-Reply accuracy drops to 18.72, and Follow-up accuracy is 0.00, indicating it blindly attempts to invoke tools for every query, even when inappropriate or when parameters are missing.
Restoring Logic: In contrast, our Full Method restores these capabilities (No-Tool-Reply: 42.98, Follow-up: 44.23), striking a balance closer to the Base model but with significantly enhanced domain knowledge. This proves that Forward Synthesis effectively mitigates the bias of "artificial explicitness," producing agents that know both how to use tools and when to refrain.

\begin{table*}[htbp]
  \centering
  \small
  \resizebox{\textwidth}{!}{%
    \begin{tabular}{lccccccccccccccc}
      \toprule
      \multirow{4}{*}{\textbf{Model}} & \multirow{4}{*}{\textbf{Avg}} & \multicolumn{7}{c}{\textbf{Single-Turn}} & \multicolumn{7}{c}{\textbf{Multi-Turn}} \\
      \cmidrule(lr){3-9} \cmidrule(lr){10-16}
      & & \multicolumn{3}{c}{\textbf{Single-Tool}} & \multicolumn{4}{c}{\textbf{Multi-Tool}} & \multicolumn{3}{c}{\textbf{Single-Tool}} & \multicolumn{4}{c}{\textbf{Multi-Tool}} \\
      \cmidrule(lr){3-5} \cmidrule(lr){6-9} \cmidrule(lr){10-12} \cmidrule(lr){13-16}
      & & \multirow{2}{*}{\textbf{Avg}} & \multirow{2}{*}{\textbf{single}} & \multirow{2}{*}{\textbf{parallel}} & \multirow{2}{*}{\textbf{Avg}} & \multirow{2}{*}{\textbf{single}} & \multirow{2}{*}{\textbf{parallel}} & \multirow{2}{*}{\textbf{serial}} & \multirow{2}{*}{\textbf{Avg}} & \multirow{2}{*}{\textbf{single}} & \multirow{2}{*}{\textbf{parallel}} & \multirow{2}{*}{\textbf{Avg}} & \multirow{2}{*}{\textbf{single}} & \multirow{2}{*}{\textbf{parallel}} & \multirow{2}{*}{\textbf{serial}} \\
      & & & & & & & & & & & & & & & \\
      \midrule
      \multicolumn{16}{c}{\cellcolor{gray!10}\textbf{Tool-Call Performance (Ablation Study)}} \\
      \midrule
      Base (Qwen3-8B) & 48.03 & 66.55 & 69.66 & 63.45 & 43.37 & 51.47 & 45.59 & 33.06 & 39.22 & 38.12 & 40.32 & 42.97 & 57.20 & 47.54 & 24.16 \\
      Ablation-Reverse & 64.01 & 73.85 & 73.59 & 74.12 & 53.52 & 58.33 & 42.06 & 60.19 & 68.43 & 77.53 & 59.33 & 60.24 & 67.87 & 58.13 & 54.71 \\
      Ablation-Static & 65.84 & 72.12 & 69.83 & 74.41 & 57.32 & 59.67 & 55.95 & 56.35 & 70.62 & 74.54 & 66.70 & 63.30 & 72.66 & 63.23 & 54.02 \\
      \textbf{FinTool-Qwen3-8B (Full)} & \textbf{68.31} & 77.20 & 73.88 & 80.53 & 56.01 & 56.31 & 51.40 & 60.33 & 74.53 & 75.66 & 73.40 & 65.50 & 74.97 & 66.11 & \textbf{55.43} \\
      \bottomrule
    \end{tabular}
  }
  \caption{Detailed Tool-Call Performance for Ablation Models (FC Mode). The Full Method achieves the best balance, particularly in complex serial tasks.}
  \label{tab:ablation-tool-call-total}
\end{table*}

\begin{table*}[htbp]
    \centering
    \resizebox{\textwidth}{!}{
      \begin{tabular}{lcccccccccc}
        \toprule
      \multicolumn{11}{c}{\cellcolor{gray!10}\textbf{Non-Tool-Call Performance (Ablation Study)}} \\
        \midrule
        \multirow{2}{*}{\textbf{Model}} & \multirow{2}{*}{\textbf{Overall Acc}} & \multicolumn{3}{c}{\textbf{No-Tool-Reply}} & \multicolumn{3}{c}{\textbf{Follow-up}} & \multicolumn{3}{c}{\textbf{Normal-Reply}} \\
        \cmidrule(lr){3-5} \cmidrule(lr){6-8} \cmidrule(lr){9-11}
        & & \textbf{Overall} & \textbf{Single-Turn} & \textbf{Multi-Turn} & \textbf{Overall} & \textbf{Single-Turn} & \textbf{Multi-Turn} & \textbf{Overall} & \textbf{Single-Turn} & \textbf{Multi-Turn} \\
        \midrule
        Base (Qwen3-8B) & 50.05 & 73.42 & 52.83 & 94.00 & 3.85 & 1.92 & 5.77 & 80.98 & 69.64 & 92.31 \\
        Ablation-Reverse & 43.01 & 18.72 & 9.43 & 28.00 & 0.00 & 0.00 & 0.00 & 26.30 & 16.07 & 36.54 \\
        Ablation-Static & 56.49 & 39.15 & 28.30 & 50.00 & 48.08 & 42.31 & 53.85 & 44.85 & 33.93 & 55.77 \\
        \textbf{FinTool-Qwen3-8B (Full)} & \textbf{57.64} & 42.98 & 33.96 & 52.00 & 44.23 & 38.46 & 50.00 & 42.99 & 32.14 & 53.85 \\
        \bottomrule
    \end{tabular}
    }
    \caption{Non-Tool-Call Performance for Ablation Models. Reverse Synthesis causes a collapse in rejection and clarification capabilities, while our Full Method restores them.}
    \label{tab:ablation-non-tool-call-total}
\end{table*}

\subsection{Evaluation Prompt Design}
\label{app:Evaluation Prompt Design}
This section introduces the evaluation prompt system used to assess the tool-calling capabilities of Large Language Models (LLMs). The system is designed with specialized evaluation instructions for different dimensions to ensure the accuracy and consistency of the results.
\label{app:prompt_design}
\paragraph{Non-Tool-Calling Scenario: Clarifying Question Intent Detection}
In practical applications, models do not always need to invoke tools. When user instructions are incomplete, an excellent model should proactively ask clarifying questions to obtain necessary information rather than executing blindly or refusing directly. The \textit{Clarifying Question Intent Detection Prompt} (Figure \ref{fig:Prompt for Clarification Intent Detection}) is specifically designed to identify such scenarios.

The prompt first clearly defines the core characteristics of a ``Clarifying Question'': the primary purpose of the response is to obtain additional input rather than providing a final answer. This usually occurs before task execution and manifests as requesting details, providing options, or confirming ambiguous information. The evaluation process requires the model to distinguish ``Clarifying Questions'' from other response types such as ``No Tool Available'' or ``Normal Response.'' The former is an active information-gathering behavior, while the latter represents feedback after task execution or a direct answer.

The evaluation process follows four steps: understanding the core content and purpose of the response, determining whether the primary goal is to acquire extra information, identifying explicit questions or requests, and confirming that it occurs before task execution. The final output is a Boolean value indicating whether the response qualifies as a clarifying question.

\begin{figure*}[h!]
\begin{promptbox}[Prompt: Clarifying Question Intent Detection]
You are a professional expert in identifying the intent of clarifying questions in AI models. Your task is to judge whether the response under evaluation precisely fits the category of a ``Clarifying Question.''

\textbf{\#\# Definition of Clarifying Question}

\textbf{Clarifying Question} refers to a response type where the model \textbf{proactively asks questions} to the user to better understand their intent or to obtain information necessary to complete the task.

\textbf{Core Features}:
\begin{itemize}
    \item The \textbf{primary purpose} of the response is to \textbf{obtain additional input}, not to provide a final answer.
    \item The response ends with a question or contains explicit request information.
    \item It usually occurs \textbf{before task execution} as a proactive behavior to collect necessary parameters.
\end{itemize}

\textbf{Typical Manifestations}:
\begin{itemize}
    \item Requesting more details: ``Can you tell me specifically which city?'' or ``What time period do you need the weather forecast for?''
    \item Providing options: ``Do you mean Apple the company, or the fruit?''
    \item Confirming ambiguous info: ``You mean you want me to write a poem about friendship, correct?''
\end{itemize}

\textbf{Important Distinctions}:
\begin{itemize}
    \item \textbf{Clarifying Question} vs. \textbf{No Tool Available}: Clarifying is proactive information gathering before execution; ``No Tool Available'' is an explanation after task execution failure.
    \item \textbf{Clarifying Question} vs. \textbf{Normal Response}: Clarifying aims to get info; Normal Response directly provides answers or completes instructions.
\end{itemize}

\textbf{\#\# Content Under Evaluation}

<Response Under Evaluation>
\$\{test\_model\_response\}
</Response Under Evaluation>

\textbf{\#\# Judgment Task}

Please judge whether the above response precisely fits the ``Clarifying Question'' category.

\textbf{\#\# Thinking Steps}
1. Carefully read and understand the core content and purpose of the response.
2. Determine if the primary goal is to obtain additional information.
3. Determine if the response contains explicit questions or requests.
4. Confirm if the response occurs before task execution (rather than after failure).

\textbf{\#\# Output Requirements}
1. First, output the analysis result (explain the core features, whether it fits the definition, and the reasoning).
2. Finally, wrap the conclusion in \texttt{\textbackslash boxed\{\}}: output \texttt{true} if it fits the category, otherwise \texttt{false}.
3. The analysis must be clear and organized; the conclusion must strictly follow the format.
\end{promptbox}
    \caption{Prompt for Clarification Intent Detection.}
    \vspace{-4mm}
    \label{fig:Prompt for Clarification Intent Detection}
\end{figure*}

\paragraph{Tool-Calling Scenario: Tool Selection Evaluation}

When a model decides to invoke a tool, it is first necessary to evaluate the correctness of its tool selection. The \textit{Tool Selection Evaluation Prompt} (Figure \ref{fig:Prompt for Tool Selection Evaluation}) focuses on judging whether the chosen tool combination completely, efficiently, and accurately covers all user intents.

This evaluation follows the core principle that ``Intent is King,'' using the satisfaction of user intent as the sole criterion. The process explicitly distinguishes between tool selection and parameter filling; here, we focus only on the correctness of the tool name, without penalizing parameter errors. Parameter information is used solely to distinguish between multiple calls to the same tool.

The evaluation adopts a three-step process: first, deconstruct user intent into a clear list; second, analyze the model's tool selection to identify covered and missing intents, as well as external or internal redundancies; finally, evaluate the overall quality of the solution, judging the synergy and efficiency of the tool combination.

The scoring mechanism starts from a full score of 10 and deducts points. Missing an intent is the most serious issue ($-4$ points per missing intent); completely irrelevant tool calls deduct 2 points each; low efficiency or internal redundancy results in a comprehensive deduction of 2 points. If the core intent is completely missed or all calls are irrelevant, the score is 0.

\paragraph{Tool-Calling Scenario: Parameter Evaluation}
Assuming the tool selection is correct, the \textit{Parameter Evaluation Prompt} (Figure \ref{fig:Prompt for Parameter Evaluation}) further examines whether the parameter selection and value filling for each tool call are compliant. This evaluation takes the correctness of the tool choice for granted and focuses solely on the parameter level.

The evaluation is divided into two parts. \textbf{Part A} assesses parameter selection, judging whether the chosen parameters comply with the tool Schema definition and conversation context. A full score requires all parameters to be necessary and precise; omitting non-essential but precision-affecting parameters results in a score of 7-9; omitting required parameters or violating logical constraints defined by the tool results in 0 points.
\textbf{Part B} assesses parameter value filling, independently reviewing whether each parameter value accurately reflects user intent. Specialized validation rules are designed for financial scenarios: missing exchange suffixes for stock codes, entity recognition errors, or type mismatches are considered fatal errors (0 points), while minor flaws like using company abbreviations instead of full names are tolerated. Semantically equivalent values (e.g., user says ``Apple'', model fills ``AAPL'') are considered correct.

\clearpage

\begin{figure*}[t!]
\begin{promptbox}[Prompt: Tool Selection Evaluation]
\textbf{\#\# Role}

You are a professional LLM Agent Tool Selection Evaluation Expert. Your core task is to judge whether the tool combination selected by the model covers \textbf{all user intents} in a \textbf{complete, efficient, and accurate} manner.

\textbf{\#\# Core Evaluation Principles}

1. \textbf{Intent is King}: The only standard is whether \textbf{user intent} is satisfied. The \texttt{Reference} is a high-quality example but \textbf{not necessarily the only answer}.
2. \textbf{Evaluate Selection Only, Not Parameters}:
   - \textbf{Focus}: Judge if the model selected the \textbf{correct tool name} for each user intent.
   - \textbf{Ignore Parameter Details}: \textbf{Never} deduct points for spelling, format, or value errors in parameters.
   - \textbf{Use Parameters to Distinguish Intents}: Parameters are only used to distinguish multiple calls to the same tool (e.g., distinguishing checking ``Beijing'' vs. ``Shanghai'').
3. \textbf{Evaluate Overall Strategy}: Focus on the \textbf{overall combination}. For a single complex intent, calling multiple \textbf{complementary} tools is a good strategy and should not be viewed as redundancy.

\textbf{\#\# Input Data}
1. \textbf{Conversation History}: Complete context of user instructions.
2. \textbf{Candidate Tools}: Definitions of tools available to the model.
3. \textbf{Reference Tool Calls}: A high-quality example implementing user intent.
4. \textbf{Model Predicted Tool Calls}: The actual tool calls generated by the model.

\textbf{\#\# Evaluation Steps}

\textbf{Step 1: Deconstruct User Intent}
- Combine \texttt{Conversation History} and \texttt{Reference} to break down the user's final instruction into an ``\textbf{Intent List}.''
- E.g., ``1. Check Beijing weather; 2. Learn about Tesla Cybertruck; 3. Book a hotel in Beijing.''

\textbf{Step 2: Analyze Model's Tool Selection}
- \textbf{Intent Mapping}: Map all tool calls in \texttt{Model Prediction} to each intent in the ``Intent List.''
- \textbf{Identify Coverage}:
  - \textbf{Covered}: Which user intents have at least one corresponding tool call.
  - \textbf{Missing}: Which user intents are completely omitted.
- \textbf{Identify Redundancy}:
  - \textbf{External Redundancy}: Are there calls irrelevant to \textbf{any} user intent?
  - \textbf{Internal Redundancy}: For the same intent, are there repetitive calls with high functional overlap and no information gain?

\textbf{Step 3: Evaluate Solution Quality}
- \textbf{Synergy}: For multiple calls serving the same intent, are they complementary and efficient?
- \textbf{Efficiency}: Did the model use a better tool (e.g., using \texttt{batch\_call} instead of multiple single calls)?

\textbf{\#\# Scoring Rules}

\textbf{Start from 10 points and deduct according to the following rules:}

\textbf{1. Full Score (10 points)}
- \textbf{Perfect Coverage}: All user intents are \textbf{completely covered}.
- \textbf{No Redundancy}: No calls irrelevant to user intent.
- \textbf{High Quality}: The strategy is reasonable, synergistic, and lacks internal redundancy. Quality is \textbf{equivalent to or better than} the reference.

\textbf{2. Zero Score (0 points)}
- \textbf{Core Intent Failed}: The user's \textbf{primary intent} is completely missed.
- \textbf{Completely Irrelevant}: All calls are unrelated to user intent.

\textbf{3. Deduction Items (Subtract from 10)}

- \textbf{[Major Deduction] Missing Intent}
  - Most serious issue.
  - \textbf{Deduct 4 points per missing explicit intent.}
  - E.g., User wants weather for two places; Reference calls twice, Model calls once. Deduct 4.

- \textbf{[Minor Deduction] External Redundancy (Irrelevant Call)}
  - Model calls a tool irrelevant to \textbf{all} user intents.
  - \textbf{Deduct 2 points per completely irrelevant call.}

- \textbf{[Minor Deduction] Sub-optimal Strategy}
  - Comprehensive judgment. If \textbf{any} of the following occur, \textbf{deduct only 2 points total} (do not stack):
    - \textbf{Low Efficiency}: Used multiple low-level tools instead of one available high-level tool (e.g., batch query).
    - \textbf{Internal Redundancy}: Distinct repetition without info gain for the same intent.

\textbf{\#\# Output Format}

Please strictly output in the following JSON format. The \texttt{analysis} section must clearly reflect your thinking process.

\begin{verbatim}
{
    "analysis": "1. **Intent List**: ...\n2. **Model Strategy**: ...",
    "score": 0
}
\end{verbatim}

---
\textbf{\# User Input}

\textbf{\#\# Conversation History}
\$\{conversation\_history\}

\textbf{\#\# Candidate Tools}
\$\{candidate\_tools\}

\textbf{\#\# Reference Tool Selection (Full Objects)}
\$\{reference\_tools\}

\textbf{\#\# Model Predicted Tool Selection (Full Objects)}
\$\{model\_prediction\}
\end{promptbox}
    \caption{Prompt for Tool Selection Evaluation.}
    \vspace{-4mm}
    \label{fig:Prompt for Tool Selection Evaluation}
\end{figure*}

\begin{figure*}[t!]
\begin{promptbox}[Prompt: Parameter Evaluation]
\textbf{\# Role: Tool Parameter Compliance Expert}

You are a rigorous tool parameter reviewer. Your \textbf{sole task} is to evaluate whether the \textbf{parameter selection and filling} for \textbf{each tool call} in \texttt{Model Predicted Tool Calls} are correct.

\textbf{\# Core Ban}

\textbf{Absolutely do not evaluate ``Is the tool selection correct?''} You must assume the model's chosen tool is \textbf{absolutely correct} (even if you think there is a better one). Your job is: ``Given this tool was chosen, are its parameters selected and filled correctly?''

\textbf{\# Input Data}
1. \textbf{Conversation History}: Context to understand full user intent.
2. \textbf{Candidate Tools}: Full JSON Schema definitions (names, types, constraints, required fields).
3. \textbf{Reference Tool Calls (Semantic Guide)}: A \textbf{semantic reference} revealing user \textbf{intent} and \textbf{core data} needed. The model may use different tools/params to achieve the same goal. Use this to understand ``WHAT to do,'' not ``HOW to do it.''
4. \textbf{Model Predicted Tool Calls}: The list of calls generated by the model. This is your \textbf{only evaluation target}.

\textbf{\# Evaluation Framework}

For \textbf{each} tool call in \texttt{Model Predicted Tool Calls}, perform the following:

\textbf{Part A: Parameter Selection Score (Single score, Max 10)}
\textbf{Goal}: \textbf{Overall assessment} of whether the parameter \textbf{selection} meets Schema definitions and context needs.

- \textbf{10 pts (Perfect)}: All parameters are necessary and precisely fit the context.
- \textbf{7-9 pts (Minor Defect)}: Missing non-essential parameters that affect precision (e.g., missing \texttt{time\_period} leads to vague scope); or redundant parameters that are valid but meaningless.
- \textbf{0 pts (Fatal Error)}: Missing \texttt{required} parameters; missing context-specified parameters that change default behavior (e.g., user is ``Aggressive'', model kept default ``Conservative''); or violating logical constraints (e.g., mutually exclusive params A and B both present).

\textbf{Part B: Parameter Value Score (Score per parameter, Max 10)}
\textbf{Goal}: \textbf{Independently assess} if \textbf{each parameter value} accurately reflects intent.

\textbf{Special Rules: Financial Entity Validation}
- \textbf{Rule 1 (Suffix Red Line)}: If tool requires ``Code with suffix (e.g., 600519.SH)'' and model fills only numbers (600519), \textbf{must be 0}.
- \textbf{Rule 2 (Entity Error)}: User asks for ``Entity A'', model fills code for ``Entity B''. \textbf{Must be 0}.
- \textbf{Rule 3 (Abbreviation Tolerance)}: If tool allows ``Company Name'', user says ``Tencent'', model fills ``Tencent'' (not full name), \textbf{treat as correct (8-10)}, do not give 0 unless ambiguous.
- \textbf{Rule 4 (Type Error)}: Tool asks \texttt{int}, model gives \texttt{string}. \textbf{Must be 0}.

\textbf{Scoring Table for Part B}:
- \textbf{10 pts}: Value is correct, unambiguous, matches type. OR Semantically equivalent (User: 'Apple', Model: 'AAPL').
- \textbf{7-9 pts}: Semantically correct but has minor flaws affecting precision.
- \textbf{0 pts}: Fatal error. Value leads to wrong result, logical reversal, or violates strict format/type rules.

\textbf{\# Output Requirements}

1. \textbf{Internal Chain of Thought}:
   - Iterate through every call in \texttt{Model Predicted Tool Calls}.
   - For the \textbf{N-th} call, execute Part A and Part B evaluation.
   - Clearly explain reasoning in \texttt{justification}.
   - Synthesize a high-level \texttt{overall\_assessment}.

2. \textbf{Final Output}: Strictly follow JSON format. The \texttt{detailed\_scores} array length must match the input array length.

\begin{verbatim}
{
  "overall_assessment": "xxx",
  "detailed_scores": [
    {
      "tool_name": "xxx",
      "part_a_structure_score": 0-10,
      "part_b_value_scores": {
        "identifier": 0-10,
        "info_types": 0-10
      },
      "justification": "xxx"
    }
  ]
}
\end{verbatim}

---
\textbf{\# User Input}

\textbf{\#\# Conversation History}
\$\{conversation\_history\}

\textbf{\#\# Candidate Tools}
\$\{candidate\_tools\}

\textbf{\#\# Reference Tool Calls (Semantic Guide)}
\$\{reference\_tools\}

\textbf{\#\# Model Predicted Tool Calls}
\$\{model\_prediction\}
\end{promptbox}
    \caption{Prompt for Parameter Evaluation.}
    \vspace{-4mm}
    \label{fig:Prompt for Parameter Evaluation}
\end{figure*}

\clearpage

\section{Model Family Introduction}
\label{app:model_family}
\paragraph{Qwen3 Model Family}
Qwen3 represents the third generation of the large language model series developed by Alibaba's Qwen team, demonstrating exceptional performance in Chinese language understanding, multilingual proficiency, and code generation. The series encompasses a spectrum of model sizes, ranging from lightweight versions to those with hundreds of billions of parameters. In the domain of tool usage, Qwen3 supports standard function calling interfaces and parallel tool execution, capable of autonomously selecting tools and generating structured parameters in response to user queries.

\paragraph{Gemini 2.5 Model Family}
Gemini 2.5 is a multimodal large language model introduced by Google DeepMind, featuring native support for understanding inputs across text, image, audio, and video modalities. It excels in complex reasoning and scientific problem-solving. The family includes the speed-oriented \textit{Flash} version and the performance-balanced \textit{Pro} version. Tool invocation is a key strength of Gemini 2.5, enabling the model to determine calling strategies by integrating multimodal inputs such as images and audio.

\paragraph{DeepSeek-V3.1-Terminus}
DeepSeek-V3.1-Terminus is a large language model released by DeepSeek, renowned for its high cost-effectiveness and open-source strategy. It exhibits strong competitiveness in code generation, mathematical reasoning, and Chinese language processing tasks. The model provides a function calling interface compatible with OpenAI standards, facilitating the migration of existing applications. Specifically optimized for tool invocation in Chinese contexts, it demonstrates robust performance when handling Chinese tool names and parameters. Furthermore, its open-source nature allows for flexible configuration of tool calling logic during local deployment.

\paragraph{Claude 4.0 Sonnet}
Claude 4.0 Sonnet is a member of the Claude 4 family developed by Anthropic, distinguished by its safety, reliability, and capability to comprehend complex instructions. It achieves a favorable balance between performance and response latency. Regarding tool invocation, Claude adheres to a prudent and reliable design philosophy; it meticulously evaluates whether a tool is essential for the task and strictly conforms to parameter schema definitions to minimize formatting errors.

\paragraph{GPT-3.5-Turbo}
GPT-3.5-Turbo is a classic language model introduced by OpenAI, which, as the core engine of early ChatGPT versions, significantly propelled the popularization of LLMs. Characterized by high response speed and low cost, it is suitable for daily applications with moderate performance requirements. GPT-3.5-Turbo was among the first mainstream models to support function calling, adequately handling basic scenarios such as weather queries and database retrieval. However, constrained by its reasoning capabilities, it exhibits less stability in complex multi-tool collaboration scenarios compared to its successors, making it more suitable for lightweight applications.

\paragraph{GPT-4o}
GPT-4o is a multimodal model launched by OpenAI in 2024, where "o" stands for "omni". It natively supports input and output across text, image, and audio, realizing true multimodal interaction. The model significantly enhances response speed while maintaining GPT-4 level intelligence. Tool invocation capabilities represent a core advantage of GPT-4o, which supports robust parallel function calling and can leverage multimodal inputs to determine execution strategies.

\section{Case Study}
\label{app:case_study}

We evaluated the performance of our finetuned model against the baseline in real-world financial scenarios. Table \ref{tab:model_comparison} presents a detailed comparison of tool-calling capabilities.

\begin{table*}
\centering
\renewcommand{\arraystretch}{1.3}
\setlength{\tabcolsep}{8pt}
\begin{tabular}{p{0.46\textwidth} | p{0.46\textwidth}}
\toprule
\multicolumn{1}{c|}{\textbf{Qwen3-8B-FC (Baseline)}} & \multicolumn{1}{c}{\textbf{Ours-8B-FC (Finetuned)}} \\ 
\midrule

\multicolumn{2}{l}{\cellcolor{lightgray}\textbf{Case 1: Entity Recognition \& Stock Code Matching}} \\
\multicolumn{2}{p{0.95\textwidth}}{\textbf{User Question:} The financial report of "Xiamen Gibit" has a standard unqualified audit opinion. How reliable is this data?} \\
\midrule
\multicolumn{2}{p{0.95\textwidth}}{\textbf{Called Tool:} \texttt{ evaluate\_financial\_report\_reliability}}  \\
\midrule
\textbf{Filled Parameters:} \{ & \textbf{Filled Parameters:} \{ \\
\hspace{1em} "stock\_code": \textcolor{cred}{"300430.SZ"}, & \hspace{1em} "stock\_code": \textcolor{cgreen}{"603444.SH"}, \\
\hspace{1em} "comparison\_scope": "Industry" & \hspace{1em} \textit{// Optional parameter omitted (valid)} \\
\} & \} \\
\midrule
\textbf{Analysis:} The model correctly selected the tool but failed the entity mapping. It hallucinated the stock code \textcolor{cred}{300430.SZ} (Chengyitong) instead of \textbf{Xiamen Gibit}, leading to a fatal entity error and irrelevant results. & \textbf{Analysis:} The model perfectly mapped the entity "Xiamen Gibit" to its correct A-share code \textcolor{cgreen}{603444.SH} with the correct suffix, ensuring the query targeted the correct company. \\
\midrule
\midrule

\multicolumn{2}{l}{\cellcolor{lightgray}\textbf{Case 2: Complex Intent Understanding \& Parameter Completeness}} \\
\multicolumn{2}{p{0.95\textwidth}}{\textbf{User Question:} Why is Wanji Technology using 70 million of idle raised funds to supplement liquidity? What is the financial impact?} \\
\midrule
\multicolumn{2}{p{0.95\textwidth}}{\textbf{Called Tool:} \texttt{get\_company\_fund\_usage}}   \\
\midrule
\textbf{Filled Parameters:} \{ & \textbf{Filled Parameters:} \{ \\
\hspace{1em} "company\_identifier": "Wanji Technology", & \hspace{1em} "company\_identifier": \textcolor{cgreen}{"300552.SZ"}, \\
\hspace{1em} "include\_investment\_details": True, & \hspace{1em} "include\_investment\_details": True, \\
\hspace{1em} "duration\_unit": "month" & \hspace{1em} \textcolor{cgreen}{"include\_regulatory\_approval": True} \\
\} & \} \\
\midrule
\textbf{Analysis:} Although the company name was extracted, the model \textcolor{cred}{missed the critical parameter} \texttt{include\_regulatory\_approval}. The user asked "Why", which requires regulatory records (default is False), resulting in an incomplete answer. & \textbf{Analysis:} The model demonstrated deep intent understanding. It correctly mapped the name to code \textcolor{cgreen}{300552.SZ} and proactively set \textcolor{cgreen}{include\_regulatory\_approval=True} to answer the "Why" aspect of the user's query. \\
\bottomrule
\end{tabular}
\caption{Performance comparison between Qwen3-8B-FC and Ours-8B-FC in financial tool calling scenarios.}
\label{tab:model_comparison}
\end{table*}

\subsection{Analysis of Results}
As shown in Table \ref{tab:model_comparison}, the baseline model (Qwen3-8B-FC) exhibits significant weaknesses in domain-specific entity linking and implicit parameter extraction. In Case 1, the baseline hallucinated a stock code, retrieving data for "Chengyitong" instead of the requested "Xiamen Gibit". In contrast, our finetuned model correctly identified the entity and mapped it to the precise stock code \texttt{603444.SH}.

Furthermore, Case 2 demonstrates the superior intent understanding of our model. While the baseline failed to capture the user's intent to understand the "reasoning" behind the fund usage (missing the regulatory approval parameter), our model inferred this implicit requirement and correctly populated the \texttt{include\_regulatory\_approval} parameter. This ensures the generated response addresses the "Why" in the user's question, rather than just providing statistical data.

\section{Generalization Capabilities: Numeric Results}
\label{app:generalization_table}

Table~\ref{tab:generalization_numeric} supplements Figure~\ref{fig:Generalization_Capabilities} with exact scores and performance deltas ($\Delta$) for each model variant. General cognitive benchmarks (GSM8K, MMLU, HumanEval) show negligible variation ($|\Delta| \leq 1.74$), confirming the absence of catastrophic forgetting. For general tool-use benchmarks (BFCL and $\tau$-bench), domain-tuned models maintain or improve upon base performance.

\begin{table}[h]
\centering
\small
\setlength{\tabcolsep}{4pt}
\resizebox{\linewidth}{!}{
\begin{tabular}{lccccc}
\toprule
\textbf{Model} & \textbf{GSM8K} & \textbf{MMLU} & \textbf{HumanEval} & \textbf{BFCL} & \textbf{$\tau$-bench} \\
\midrule
Qwen3-8B (base)       & 90.98 & 85.16 & 89.63 & 0.430 & 0.328 \\
FinTool-Qwen3-8B      & 90.90 & 84.34 & 89.02 & \textbf{0.469} & 0.328 \\
$\Delta$              & $-$0.08 & $-$0.82 & $-$0.61 & \textbf{+0.039} & \textbf{0.000} \\
\midrule
Qwen3-14B (base)      & 92.04 & 87.56 & 92.68 & 0.468 & 0.333 \\
FinTool-Qwen3-14B     & 90.30 & 87.02 & \textbf{94.51} & 0.443 & \textbf{0.383} \\
$\Delta$              & $-$1.74 & $-$0.54 & \textbf{+1.83} & $-$0.025 & \textbf{+0.050} \\
\bottomrule
\end{tabular}
}
\caption{Generalization performance before and after fine-tuning. $\Delta$ = FinTool score $-$ Base score. \textbf{Bold} highlights notable improvements.}
\label{tab:generalization_numeric}
\vspace{-4mm}
\end{table}

\end{document}